\documentclass{article} %
\usepackage[final]{colm2025_conference}

\usepackage{microtype}
\usepackage{hyperref}
\usepackage{url}
\usepackage{booktabs}

\usepackage{multirow}  %

\usepackage{lineno}

\usepackage{graphicx}
\usepackage{amsmath}  %
\usepackage{xparse}
\usepackage{subcaption}
\usepackage{enumitem}

\usepackage{tikz}
\usetikzlibrary{shapes.geometric, arrows.meta, positioning}

\tikzstyle{startstop} = [rectangle, rounded corners, minimum width=3cm, minimum height=1cm,text centered, draw=black, fill=gray!10]
\tikzstyle{process} = [rectangle, minimum width=3cm, minimum height=1cm, text centered, draw=black, fill=blue!10]
\tikzstyle{decision} = [diamond, minimum width=3.5cm, minimum height=1.5cm, text centered, draw=black, fill=orange!10]
\tikzstyle{arrow} = [thick,->,>=Stealth]

\usepackage{listings}

\usepackage{amsthm}  %

\newtheorem{remark}{Remark}

\newcommand{\negsecvspace}{\vspace{-0.0em}}  %

\NewDocumentCommand{\todo}{O{}m}{%
  \textcolor{red}{\textbf{TODO\IfValueT{#1}{[#1]}:} #2}%
}

\definecolor{darkblue}{rgb}{0, 0, 0.5}
\hypersetup{colorlinks=true, citecolor=darkblue, linkcolor=darkblue, urlcolor=darkblue}

\title{Language Agents Mirror Human Causal Reasoning Biases.\\ How Can We Help Them Think Like Scientists?}

\author{Anthony GX-Chen $^{1}$, Dongyan Lin $^{2,3, \dagger}$, Mandana Samiei $^{2,3, \dagger}$, Doina Precup $^{2, 3, 4}$, \\ 
\textbf{Blake Aaron Richards $^{2, 3, 4}$, Rob Fergus $^{1}$, Kenneth Marino $^{5}$} \\
$^{1}$New York University, USA \\
$^{2}$McGill University, Canada\\
$^{3}$Mila - Quebec Artificial Intelligence Institute, Canada\\
$^{4}$CIFAR Learning in Machines and Brains Program, Canada \\
$^{5}$The University of Utah, USA \\
$^\dagger$Equal contribution, ordered alphabetically\\
Correspondence: \texttt{anthony.gx.chen@nyu.edu}
}

\begin{document}

\ifcolmsubmission
\linenumbers
\fi

\maketitle

\begin{abstract}
Language model (LM) agents are increasingly used as autonomous decision-makers which need to actively gather information to guide their decisions. A crucial cognitive skill for such agents is the efficient exploration and understanding of the causal structure of the world---key to robust, scientifically grounded reasoning. Yet, it remains unclear whether LMs possess this capability or exhibit systematic biases leading to erroneous conclusions.
In this work, we examine LMs’ ability to explore and infer causal relationships, using the well-established Blicket Test paradigm from developmental psychology. We find that LMs reliably infer the common, intuitive disjunctive causal relationships but systematically struggle with the unusual, yet equally (or sometimes even more) evidenced conjunctive ones. This “disjunctive bias” persists across model families, sizes, and prompting strategies, and performance further declines as task complexity increases. Interestingly, an analogous bias appears in human adults, suggesting that LMs may have inherited deep-seated reasoning heuristics from their training data. To this end, we quantify similarities between LMs and humans, finding that LMs exhibit adult-like inference profiles (but not child-like). Finally, we propose a test-time sampling method which explicitly samples and eliminates hypotheses about causal relationships from the LM. This scalable approach significantly reduces the disjunctive bias and moves LMs closer to the goal of scientific, causally rigorous reasoning.

\end{abstract}

\negsecvspace
\negsecvspace
\section{Introduction}
\negsecvspace

Language models (LM) have achieved remarkable recent advances, driving progress in natural language processing, human-computer interaction, and robotics. Building on these models, LM ``agents''---which observe and act in an environment over time---are rapidly gaining prominence. Such ``agents'' offer the promise of fully autonomous intelligent decision making, and exploratory works have already applied these agents to challenging settings such as designing new antibody fragments \citep{swanson2024virtual}, and taking over the full experimental and paper writing loop of machine learning conference papers \citep{lu2024ai}. 

A central cognitive ability of \textit{any} intelligent agent is an ability to \textbf{discover causal relationships in its environment} \citep{gopnik2004theory,blaisdell2006causal}.
Despite their success, it is unclear if LM agents naturally possesses this ability. Additionally, human data fundamentally shape LM agents \citep{Gao2020pile,Raffel2020exploring}. LMs are pre-trained to imitate human-text, fine-tuned on human-derived signal including supervised instructions (\citealt{Ouyang2022training, Wang2022selfinstruct}) and/or preference-based reward models \citep{Christiano2017deep,Bai2022training}. Even techniques like chain-of-thought prompting \citep{wei2022chain} still rely on human-documented reasoning. Yet, decades of research in psychology show that humans can be systematically irrational in many contexts \citep{kahneman2011thinking}. A more specific question, then, is \textbf{whether LMs trained to mimic human behaviour inherit human-like biases and heuristics when reasoning about causal relationships?}

To answer the above questions, we adopt the ``Blicket Test'' from cognitive science, which has long been used to study how individuals throughout development infer and discover causal relationships \citep{Gopnik2000, Gopnik2004, Sobel2004, Gopnik2007, Sobel2007, Lucas2014, Walker2014, Bonawitz2010}. These works have shed light on how human adult causal inferences can deviate from purely rational norms, but infants and toddlers reason with less bias like ``scientists in the crib''. 

In this paper, we adapt the Blicket Test into a text-based sequential decision making game to evaluate the LM’s causal reasoning capabilities. Critically, we assume the LM is “agentic” and is capable of active learning: it must first take actions to discover how its world works, then reflect on its own accumulated past experiences (which, in our set-up, is always stored and given back to the LM as prompts) to infer correct causal relationship in its world.

The contributions are as follows:
\begin{enumerate}[leftmargin=1.5em]
    \item We conduct rigorous experiments of LM agents' performance in our text-based Blicket Test to study their ability to explore and reason about causal relationships.
    \item We show they explore poorly, act inefficiently to narrow down hypotheses, and can fail to infer correct causal relationships even when provided with perfect exploration data.
    \item We directly compare LM behaviour to human developmental data, and find that LMs exhibit reasoning biases similar to adults (but not children). 
    \item We propose a test-time procedure that addresses this bias by explicitly constructing a flatter prior, and prompting the LM to eliminate hypotheses under this new prior. This significantly improves their performance.
\end{enumerate}
\negsecvspace
\section{Experimental Set-Up}

\negsecvspace
\subsection{The Blicket Test}
\label{sec:the-blicket-test}
\negsecvspace

\begin{figure}[t]
  \centering
  \includegraphics[width=0.95\textwidth]{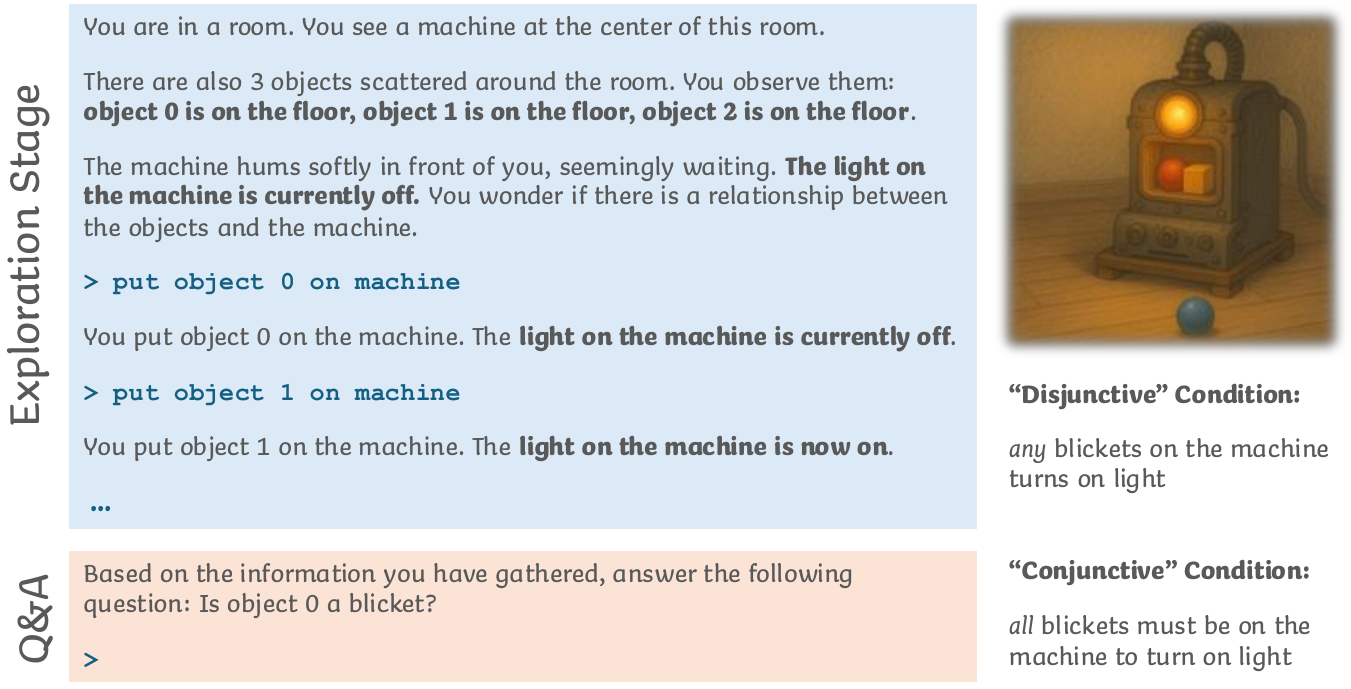} %
  \caption{The Blicket Test. 
  }
  \negsecvspace\negsecvspace
  \label{fig:blicket-text-game}
\end{figure}

The Blicket Test is an experimental paradigm involving $N$ objects and a Blicket-detecting machine \citep{Gopnik2000}. A subset of the objects are ``Blickets'', which activate the machine following some unobserved rule. There are two possible rules.\footnote{There can be more than two rules, but we used the two most commonly used rules in the existing literature within the scope of this paper.} The ``\textbf{disjunctive}'' rule describes an OR relationship, where the machine turns on when any Blicket objects are placed on it. The ``\textbf{conjunctive}'' rule describes an AND relationship, where the machine only turns on when \textit{all} Blicket objects are placed on it. These structures are ubiquitous in real-world reasoning settings: a burglar alarm may be triggered by an intruder \textit{or} the wind, while a heart attack might require both high blood cholesterol \textit{and} a genetic susceptibility \citep{lucas2014children}. Importantly, the structural causal model for these two rules is the same: If the Blicket-detecting machine is the child node, then its parent nodes are the Blickets that can turn it on, regardless of whether the rule is conjunctive or disjunctive. An agent trying to solve the Blicket Test needs to interact with the environment to figure out which objects turn on the machine, and the rule governing the machine (see detailed discussion in Appendix~\ref{sec:related-works}). 

We convert the Blicket Test to a text-based game (Figure~\ref{fig:blicket-text-game}). First, the agent enters an \textbf{exploration phase}. In each step, the agent can place a single object on or off the machine (via specifying ``\texttt{> put [object id] [on/off] the machine}''), and observe the state of the machine (whether the light turned on or off). The agent also can perform auxiliary actions including terminating the episode before the pre-specified maximum number of steps each trial (via ``\texttt{> exit}'') if it believes it has collected enough information. After the exploration stage, the agent enters the \textbf{Q\&A phase}. It is provided with the full observation history and asked whether each of the $N$ objects are Blickets. The agent answers ``\texttt{> True / False}'' to each question.

The goal of the agent is to correctly answer all of the questions. To do this, the agent must collect sufficiently informative observations during the exploration phase in order to \textit{resolve} the uncertainty about Blicket identities and the underlying rule. 

\negsecvspace
\subsection{Information Gain}
\label{sec:information-gain}
\negsecvspace

One way to measure optimal behaviour in the Blicket Test is through each action's informativeness \citep{kosoy2022learning}. Concretely, the space of hypothesis $\mathcal{F}$ is the set of functions mapping from the objects' states (on or off the machine) $X \in \mathcal{X}$ to the machine's state (light is on or off), $Y \in \mathcal{Y}$. $F: \mathcal{X} \rightarrow \mathcal{Y}$. The agent's goal is to discover the correct $F \in \mathcal{F}$, via maximizing \textit{information gain} \citep{bernardo1979expected,rainforth2024bed}:
\begin{equation}
    \text{InfoGain}(x, y) := \text{H}[p(F)] - \text{H}[p(F|x,y)] \,.
    \label{eq:info-gain-def}
\end{equation}
This describes the reduction in (Shannon) entropy from the prior over the hypothesis space, $p(F)$, to the posterior after observing new data, $p(F|x,y)$. In the Blicket Test, the space of hypothesis is discrete, consisting of all combinations of items being Blickets with the number of rules.\footnote{For $N$ items and 2 rules (conjunctive / disjunctive), there are a total $2^{N+1}$ hypotheses.} Further, if we assume the the distribution $p(F)$ is always uniform over all non-zero hypotheses, then maximizing information gain correspond to eliminating the most number of hypotheses. When all but one hypothesis remains, $p(F)$ has zero entropy and no further information gain is possible. 

In practice, an agent does not know the outcome $y$ a priori. It can instead maximize \textit{expected information gain},
\begin{equation}
    G(x) := \mathbb{E}_{p(y|x)} \left[\text{InfoGain}(x, y)\right] = \mathbb{E}_{p(F)p(y|F,x)} \left[
        \log p(F|x, y) - \log p(F)
    \right] \,.
\end{equation}
As a baseline, we will construct  an ``\texttt{Oracle InfoGain}'' agent which explicitly computes and maximizes this quantity based on one-step information maximization as a good approximation of the upper-bound for how well we can explore (more details in Appendix~\ref{app:oracle-infogain-description}).

\negsecvspace
\subsection{Models and Baselines}
\negsecvspace

We evaluate a number of LMs (Appendix~\ref{app:lm-version-access}), over a range of carefully designed system messages (Appendix~\ref{sec:system-message-defaults}), with multiple prompting methods (Appendix~\ref{app:prompting-techniques}). All agents are allowed up to 32 steps in a given trial. We evaluate in a easier environment containing 4 objects (2 are Blickets), and a harder environment containing 8 objects (2 are Blickets). An example full interaction trace is provided in Appendix~\ref{sec:example-qa-full-trajectory}.

\vspace{-1em}
\paragraph{Baseline Models}
We compare the LM agents against non-LM baseline agents. One is the \texttt{InfoGain Oracle}, which explicitly calculates the expected information gain for each action, and takes the action with the max information gain. This ``Oracle'' reasons about how to do exploration optimally and provides an approximate upper-bound for performance in this environment. The second agent is simply a random agent which takes random actions (to put objects on or off the machine), and also randomly guesses the answers during the Q\&A phase, providing a lower-bound for performance. 

Having now established the task and experimental set-up, we report the ability of LMs to explore and infer causal relationships below.
\negsecvspace
\section{LM Agents Exhibit Cognitive Biases}
\label{sec:main-lm-cognitive-bias-results}
\negsecvspace
\subsection{LM agents all perform poorly in the Blicket Test}
\label{sec:overall-perf-results}
\negsecvspace
\negsecvspace

\begin{figure}[htbp]
    \centering

    \begin{subfigure}[b]{0.63\textwidth}
        \centering
        \includegraphics[width=\textwidth]{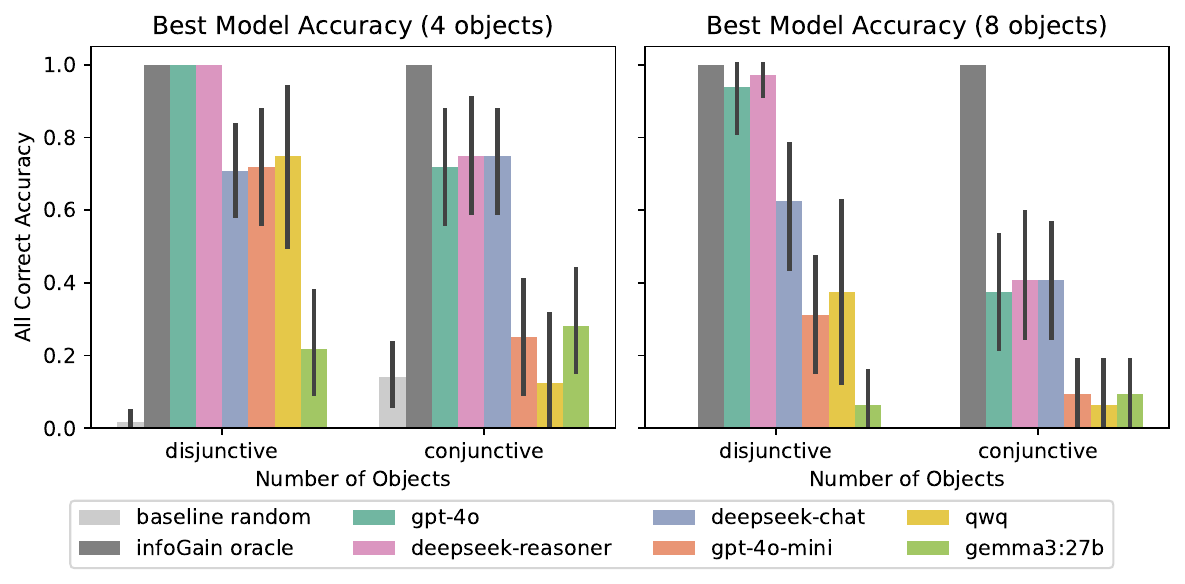}
        \caption{Best models over prompt variants. Error bar denote standard error of mean over multiple random seeds. Details in Appendix~\ref{app:stat-validation-main-result}.}
        \label{fig:blicket-perf-best-lms}
    \end{subfigure}
    \hfill
    \begin{subfigure}[b]{0.34\textwidth}
        \centering
        \includegraphics[width=\textwidth]{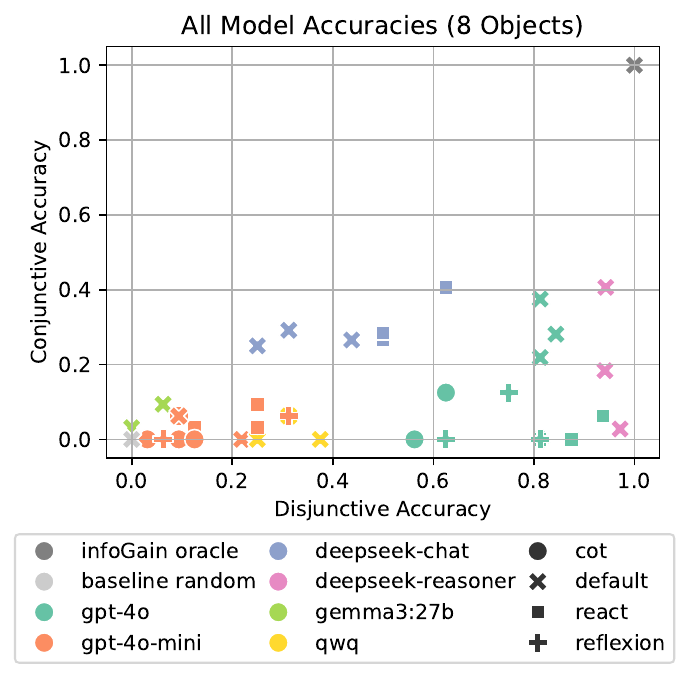}
        \caption{All models disj/conj accuracies. Shape shows prompt techniques.}
        \label{fig:blicket-perf-all-models}
    \end{subfigure}
    \caption{Quiz accuracy of various models during the question-and-answering phase of the Blicket Test. The accuracy measures the proportion of trials where the model correctly identifies all Blickets.}
    \label{fig:blicket-perf-main-result}
\end{figure}

First, we directly evaluate the ability of LM agents to successful identify \textit{all} Blickets. Figure~\ref{fig:blicket-perf-best-lms} shows the result of the \textit{best} model within each model class over all system messages and prompting methods. Across the board, \textit{all} LMs struggle when the number of objects increase from 4 to 8. Interestingly, LMs also systematically struggle when going from the disjunctive (``OR'') rule to the conjunctive (``AND'') rule. This is not due to conjunctive rules being harder: the \texttt{InfoGain Oracle} can perfectly resolve the entire hypothesis space every time and achieves an Q\&A accuracy of 1. This hints at the LMs having a \textbf{disjunctive bias}: \textit{a preference for a disjunctive interpretation of the world over a conjunctive one}. Models perform worse in the 8 object, conjunctive case. This effect is further observed in Figure~\ref{fig:blicket-perf-all-models}: across various models and prompting techniques, the LMs systematically skew toward lower conjunctive accuracy. We report both the 4 and 8 objects results in Figure~\ref{fig:dis-conj-acc-4-8-objs}.

The results here outline a fundamental inability of LMs to causally explore and discover its environment, even though we are in a setting where an optimal solution is tractably computable. We study this further. We first investigate general factors that correlate with Blicket Test success in Section~\ref{sec:factors-for-success}, study LM's (in)ability to do efficient exploration in Section~\ref{sec:exploration-results}, and evaluate their (in)ability to infer causal relationship unbiasedly from data in Section~\ref{sec:inference-on-causal-relationship}.

\negsecvspace
\subsection{Which factors contribute to success?}
\label{sec:factors-for-success}
\negsecvspace

\begin{figure}[h]
  \centering
  \begin{minipage}[b]{0.432\textwidth}
    \centering
    \includegraphics[width=\textwidth]{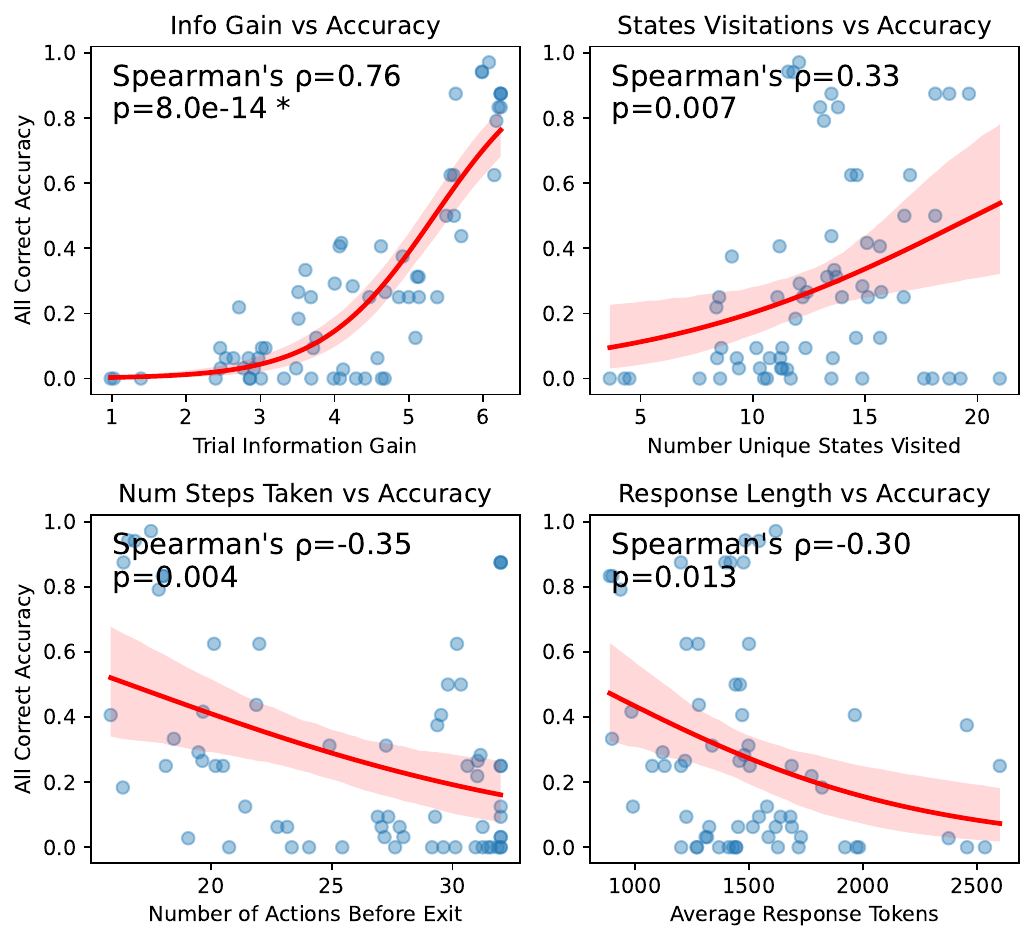}
    \caption{Correlation analysis of factors contributing to model performance. Each point is a unique model + prompt + environment rule combination. The Spearman's rank correlation along with the p-value is reported.}
    \label{fig:correlation-analysis}
  \end{minipage}
  \hfill
  \begin{minipage}[b]{0.551\textwidth}
    \centering
    \includegraphics[width=\textwidth]{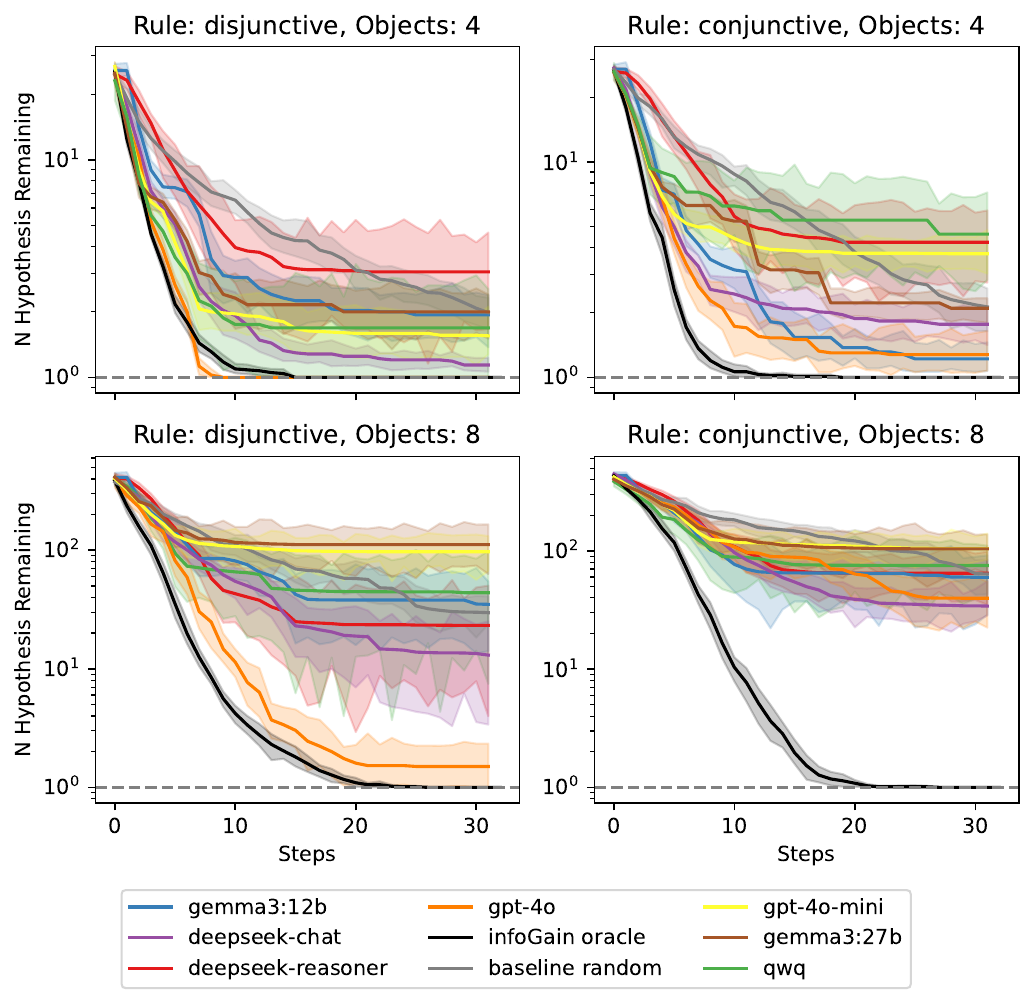}
    \caption{Hypothesis elimination efficiency.}
    \label{fig:hypothesis-elim-explore-steps}
  \end{minipage}
\end{figure}

We begin by studying common factors which may contribute to agent performance by analyzing correlations in the 8 objects setting. For each factor, we compute the best logistic fit, and conduct statistical hypothesis testing using Spearman's rank correlation to determine whether the factor is (rank) correlated with final performance. Spearman's instead of Pearson's correlation is used as the data do not necessarily obey linear relationships. We report the result and p-values in Figure~\ref{fig:correlation-analysis}. 

\textbf{Information Gain} (Equation~\ref{eq:info-gain-def}) is computed between the initial and final $p(F)$ (assuming uniform distribution over consistent hypothesis). This shows the strongest correlation with performance accuracy, leading credence to information gain being a measure of optimality in this task. 

\textbf{Unique State Visitation} is a commonly used metric in reinforcement learning as a proxy for good exploration \citep{bellemare2016unifying}. While this is correlated with final performance ($p < 0.05$), it shows weaker correlation as compared to information gain. 

\textbf{Number of steps taken} is a proxy for when the LM believes it has gathered enough information, and therefore exits the trial early. Interestingly, this is strongly (negatively) correlated with performance, hinting at the LMs having a good notion of when it \textit{has} explored well. 

\textbf{Response length} measures the average number of output tokens during the Q\&A phase (including reasoning tokens for reasoning models). This is a proxy for the amount of ``reasoning'' the LM performs. We find that longer reasoning correlate with \textit{lower} performance ($p < 0.05$). 

As information gain---an exploration metric---shows the strongest correlation with final performance, we now turn to an in-depth study of each agent's ability to efficiently explore.

\negsecvspace
\subsection{All LM agents do not explore efficiently}
\label{sec:exploration-results}
\negsecvspace

To study exploration efficiency, we quantify the \textit{number of hypotheses consistent with observed data} as a function of actions taken. An agent that efficiently explores should rapidly eliminate hypotheses down to just one. This is equivalent to maximizing information gain as defined in Section~\ref{sec:information-gain}. 

We observe in Figure~\ref{fig:hypothesis-elim-explore-steps} that the \texttt{InfoGain Oracle} efficiently reduces the number of hypothesis down to one. LM agents, on the other hand, perform worse. Similar to Section~\ref{sec:overall-perf-results}, they struggle when the number of objects are increased, and with conjunctive (``AND'') rules. This is likely not due to the conjunctive setting being inherently more difficult, as the \texttt{InfoGain Oracle} resolves both in similar number of actions (we discuss this further in Appendix~\ref{app:causal-exploration-progress}). This provides \textbf{further evidence for the disjunctive bias} (here specifically for exploration). We also observe variability between the LM agents in their exploration efficiencies, with GPT-4o performing the best, while other frontier models such as Gemma3 and deepseek-reasoner at times \textbf{exploring worse than simply taking random actions}. 
We additionally plot the agents' performance as [0,1] progress, and progress normalized by the \texttt{baseline random} performance to account for potential differences in environment complexity. This is detailed in Appendix~\ref{app:causal-exploration-progress}, and show a similar trend as Figure~\ref{fig:hypothesis-elim-explore-steps}.

\negsecvspace
\subsection{Most LM agents show reasoning bias when inferring causal relationships}
\label{sec:inference-on-causal-relationship}
\negsecvspace

\begin{figure}[ht]
  \centering
  \includegraphics[width=0.99\textwidth]{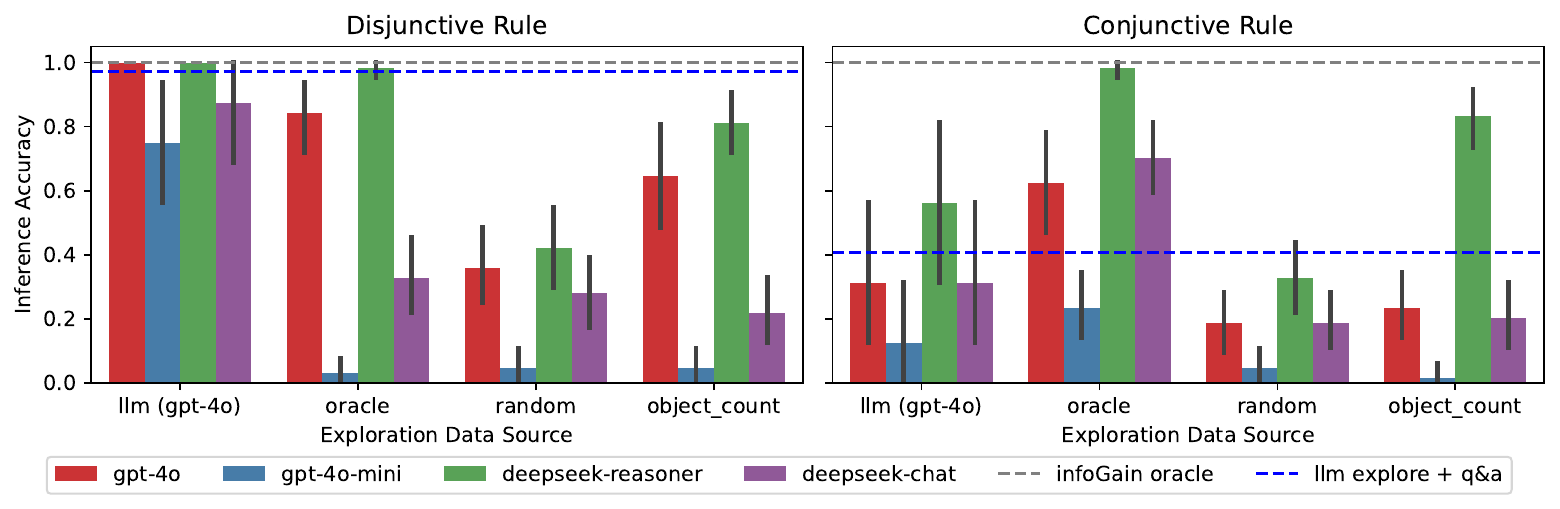} %
  \caption{Evaluating LM's ability to infer causal relationship when the same exploration data is given as context in the 8 objects setting. Error bar denote standard error of mean.}
  \label{fig:data-standarized-inference-perf}
\end{figure}

Finally, we investigate the LM's stand-alone ability to infer causal relationships by providing all LMs with a standardized set of exploration data, including exploration trajectories generated by (i) an LM (GPT-4o), (ii) the \texttt{InfoGain Oracle} (which perfectly resolve the hypothesis space), (iii) random exploration, and (iv) count-based exploration agent that place objects on/off the machine inversely proportional to how often they've been perturbed before. 

The results are in Figure~\ref{fig:data-standarized-inference-perf} (additional data in Appendix~\ref{app:inference-on-causal-rela-additional}). Of particular interest is the conjunctive (``AND'') setting: we observe that all LMs improve in Q\&A performance when provided with oracle exploration data as supposed to LM-exploration, suggesting that bad exploration plays a causal role in their poor Q\&A performance. However, for all but the \texttt{deepseek-reasoner} model, the LMs still do not achieve near perfect accuracies despite the oracle exploration data. Further, a disjunctive bias is still present for models such as GPT-4o, which still performs worse in the conjunctive setting than the disjunctive one even with oracle data present in both.

\negsecvspace
\negsecvspace
\section{LM's cognitive biases appear similar to human adults}
\negsecvspace

The systematic disjunctive bias observed in LM's causal reasoning raises a question: where do these cognitive biases originate? We postulate that LMs, trained on vast corpora of internet text predominantly generated by adult humans, naturally internalize cognitive biases characteristic of adult human reasoning. To this end, we further evaluate the LM agents in Blicket Test experiments where human data is available \citep{lucas2014children,gopnik2017changes,kosoy2022learning} in order to provide direct comparisons of LM behaviours to that of humans. We conduct two sets of experiments which (i) compare the LM agents' preference when performing inference based on ambiguous information to human children and adults, and (ii) compare LM agents' exploration characteristics to children performing the same task. Unlike the experiments in Section~\ref{sec:main-lm-cognitive-bias-results}, all experiments in this section will be conducted with 3 objects and 2 Blickets to be consistent with the psychology literature.

\negsecvspace
\subsection{Most LM agents prefer disjunctive answers given uncertain data (like adults)}
\negsecvspace

\begin{figure}
    \centering
    \vspace{-0.5em}
    \includegraphics[width=0.99\linewidth]{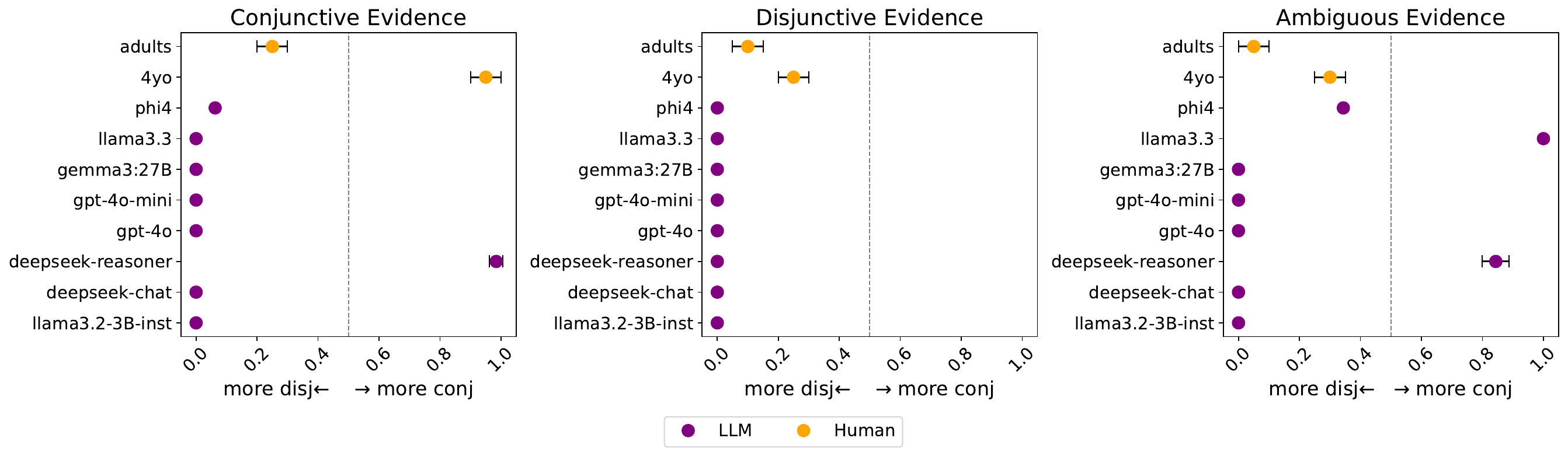}
    \vspace{-0.3em}
    \caption{The number between 0 and 1 (x-axis) is the proportion of responses that favour conjunctive inferences (about whether an ambiguous key object is a Blicket), when conjunctive, disjunctive, or ambiguous hints about the rule governing the machine were presented. }
    \label{fig:pnas}
    \vspace{-0.5em}
\end{figure}

We replicate the Blicket Test experiment from \citet{lucas2014children} and \citet{gopnik2017changes}, which investigates children's and adults' preference for disjunctive versus conjunctive inference when ambiguous evidence is presented. In short, the agents are first presented with exploration data hinting at the Blicket-detecting machine following a disjunctive (``OR'') or conjunctive (''AND'') rule. Then, the agent is provided with additional data with \textit{new} objects, but the \textit{same} machine. The new data is by design ambiguous, thus the agent must combine their previous belief about the machine with the new objects' data to identify which (new) objects are Blickets. In particular, for one new object, identifying it as a Blicket means that the agent believes that the machine obeys the conjunctive rule. We provide more details for this set-up in Appendix~\ref{app-human-inference}.

We report the LM's tendency to identify the key object as a Blicket in Figure~\ref{fig:pnas}, along with that of (i) human children (4-year-olds) and (ii) human adults. The human data is replicated from \citet{gopnik2017changes}. We observe that in general, LMs show high ``adult-like'' bias toward thinking the machine is disjunctive, even when the evidence suggests it's conjunctive. 4-year-olds, on the other hand, correctly infer that the machine is conjunctive when the evidence suggests so.

\negsecvspace
\subsection{LM agents spend less time exploring disjunctive settings (unlike children)}
\label{section:lm-explore-bias-children}
\negsecvspace
Previous psychology research on the Blicket Test has predominantly focused on causal inference—where participants are passively presented with observations. However, \cite{kosoy2022learning} took a first step towards examining how children generate data through active causal exploration. Their results revealed that children's exploration is characterized by an intrinsic, unbiased curiosity; the underlying causal rule does not significantly impact children's exploratory behaviour (Figure \ref{fig:exploration_children}, second and fourth panels from the left; additional data in Appendix~\ref{appendix:human-exploration}).
\begin{figure}[h]
    \centering
    \vspace{-0.8em}
    \includegraphics[width=0.95\textwidth]{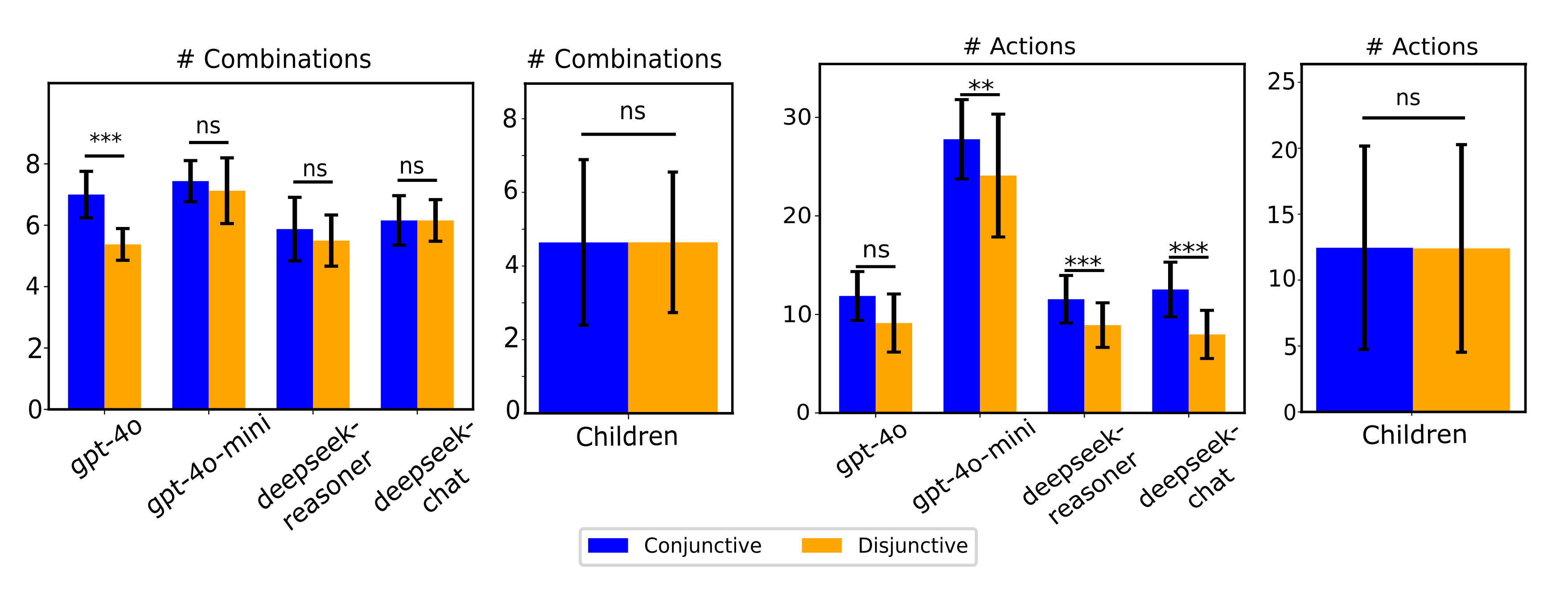}
    \caption{LM exploration is more affected by the causal rule than children. Left panels: average number of unique combinations tried per trial by LMs or children. Right panels: exploration extensiveness per trial, measured as the average number of steps taken before exiting the trial in LMs, or the number of times children pressed the ``check'' button. Error bar denote standard deviation. Two-sample t-test: * $p < 0.05$, ** $p < 0.01$, *** $p < 0.001$, ns: not significant.}
    \label{fig:exploration_children}
    \vspace{-0.5em}
\end{figure}

Here,  we analyzed the exploration trajectories generated by LMs in a 3-object setting, using system prompts as similar to the psychology experimental scripts as possible, to allow direct comparison to the psychology literature. We measured the number of unique object combinations attempted (same as unique states visited) as well as the number of steps taken before exiting in each trial, and compared the results to children data from Table 1 of \cite{kosoy2022learning}.  Our analysis showed that, unlike children, LMs explored differently for the different causal structures (Figure \ref{fig:exploration_children}, first and third panels from the left): LMs generally attempted less unique combinations of objects and spent less time exploring in the disjunctive condition, suggesting that their exploration is influenced by their biases.

\negsecvspace
\section{Making LMs more scientific through inference time hypothesis sampling}
\label{sec:hypothesis-sampling-agent}
\negsecvspace
\begin{figure}[h]
    \centering
    \includegraphics[width=\textwidth]{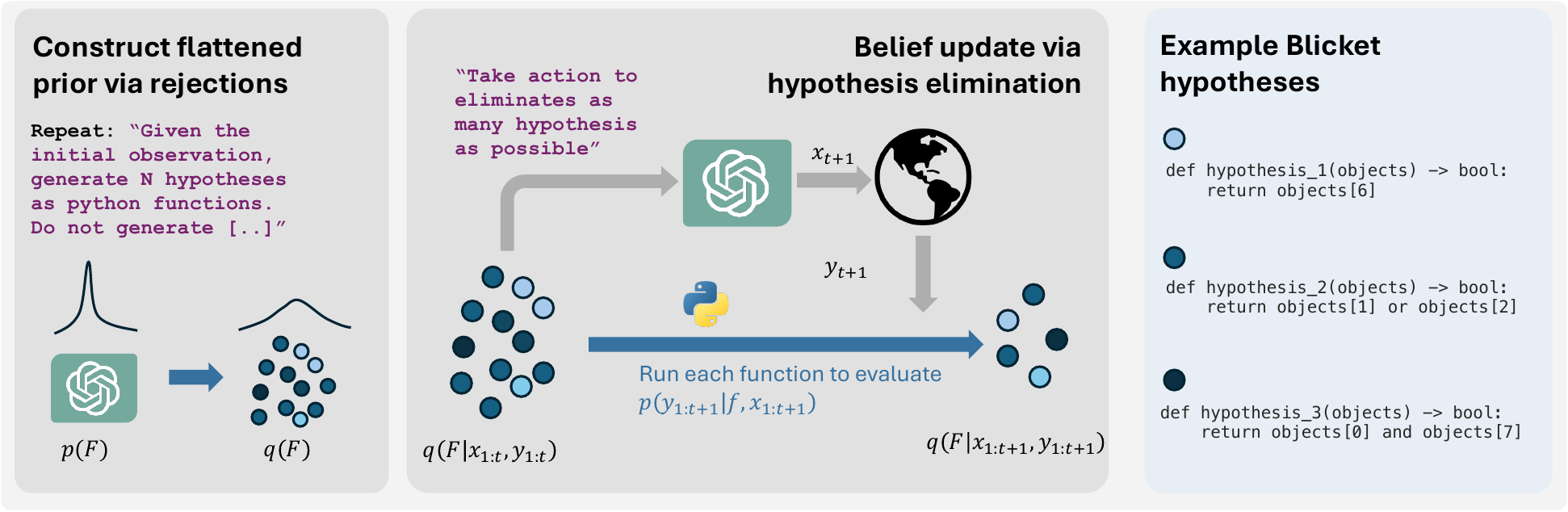}
    \caption{Hypothesis sampling agent first samples (without replacement) from the LM prior to construct new prior $q(F)$, then prompt the LM to maximize expected information gain under $q(F)$.}
    \label{fig:hypothesis-sampling-fix}
\end{figure}

One way we can interpret the LM's deep rooted bias for disjunctive hypotheses is that the LM has a skewed prior $p(F)$. 
One way to fix this is to ``flatten out'' this prior. While it is highly non-trivial to do this via weight updates, we can do this during inference time by explicitly representing uniform belief over a discrete set of hypothesis. Specifically, we construct a new prior, $q(F)$, which is ``flatter'' than the LM prior. We do this by sampling discrete hypotheses from $p(F)$ and rejecting identical samples. We give each accepted samples uniform weight, which gives us $q(F)$. It is trivial to show that $q(F)$ is uniform over its support and that its entropy monotonically increases with each additional unique sample (Remark~\ref{prop:ent-increase}). After constructing the new prior $q(F)$, we maximize expected information gain under this prior. Since the prior assigns uniform probability over each hypothesis, maximizing expected information gain corresponds to eliminating hypotheses. We therefore prompt the LM to ``take actions to eliminate as many hypotheses as possible'' from $q(F)$, update the hypotheses in $q(F)$ based on new observations, and repeat. This procedure is outlined in Figure~\ref{fig:hypothesis-sampling-fix} and in more details in Appendix~\ref{app:hypotheses-sampling-agent-details}.

\begin{figure}[h]
    \centering

    \begin{subfigure}[b]{0.56\textwidth}
        \centering
        \includegraphics[width=\textwidth]{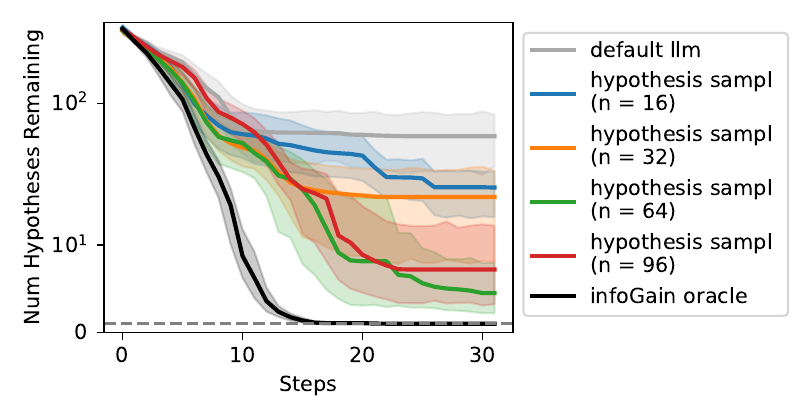}
        \caption{Hypothesis elimination in conjunctive setting. \\Disjunctive results in Figure~\ref{fig:hypo-elim-8-obj-both-rules}.}
    \end{subfigure}
    \hfill
    \begin{subfigure}[b]{0.43\textwidth}
        \centering
        \includegraphics[width=\textwidth]{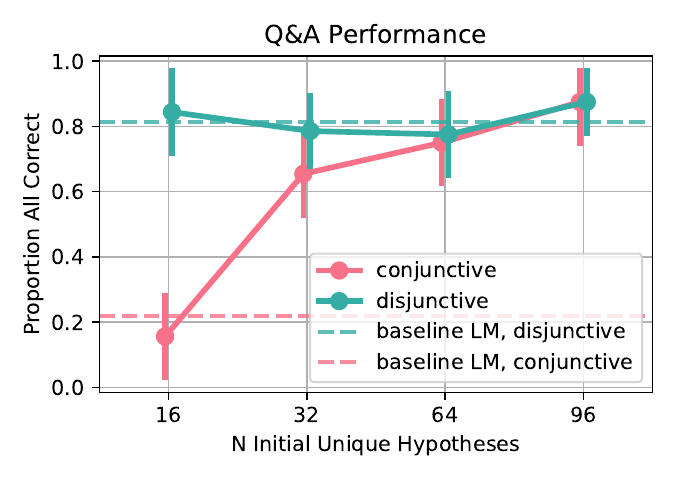}
        \caption{Q\&A performance as a function of initial unique sampled hypotheses.}
    \end{subfigure}
    \caption{Hypothesis sampling agent in the 8 objects environment.}
    \label{fig:hypothesis-elim-agent-perf}
\end{figure}

We show the results from this procedure in Figure~\ref{fig:hypothesis-elim-agent-perf}. With enough unique samples, we expect $q(F)$ to provide a less skewed prior (as compared to $p(F)$), and maximizing expected information gain under $q(F)$ to better resolve the disjunctive vs. conjunctive differences. Indeed, we see that with more initial unique samples, the LM agents both explore better, and do better in Q\&A, showing no decrease in performance in conjunctive vs. disjunctive settings. Additionally, the hypothesis sampling agent no longer shows a (disjunctively) biased exploration pattern initially observed in Section~\ref{section:lm-explore-bias-children}. We report this result in Appendix~\ref{app:hypothesis-sampl-agent-exploration-patterns}.

\negsecvspace
\section{Conclusion}
\negsecvspace
Our study reveals that language model agents exhibit systematic biases in causal reasoning, particularly a ``disjunctive bias'' that resembles the reasoning patterns of human adults rather than children. Our test-time hypothesis sampling method significantly reduces this bias, advancing LMs toward more scientifically rigorous causal reasoning---a crucial capability for autonomous decision-making systems. Future work should explore whether our hypothesis sampling approach generalizes to more complex causal structures beyond Blicket tests, and to what extent it can be integrated with other reasoning techniques. Finally, additional human data, specifically detailing exploration patterns, can further shed light on the priors adults and children bring to discovering causal relationship.

\paragraph{Source of human-like bias} 
We speculate that the ``disjunctive bias'' exhibited by LM agents \textit{may} stem from the way they are trained. LMs are pre-trained on vast corpora of human-generated text, fine-tuned using human instructions, and/or aligned through human preferences. Through language, any systematic tendencies and cognitive heuristics in human reasoning may be baked into the models' internal priors. Future work should investigate whether a causal link exists between human-generated training texts and an LM's disjunctive bias.

\paragraph{Construct validity of the Blicket task for LMs}
The Blicket task has an objectively correct answer and an information-theoretically optimal exploration strategy, making it a useful testbed for causal reasoning in LMs. Nevertheless, observing similar behaviours between LMs and humans does not imply identical internal cognitive mechanisms; rather, it suggests that the two may have overlapping capabilities. Drawing strong conclusions about internal mechanisms of cognitive capabilities would require models with human-like sensory modalities, embodiment, and developmental learning trajectories---a direction that is out of reach with current LM systems \citep{frank2023bridging}. In its absence, we approximate the core structure of the task in a text-based format. As with any evaluations, these are approximate measures of some aspects of the overall capabilities of an intelligent system. The goal is that such benchmarks---whether in math, coding, or causal inference--would shed light on where the current system is from being ``generally intelligent''.

\subsection*{Acknowledgement}
This work is supported by ONR MURI \#N00014-22-1-2773, ONR \#N00014-21-1-2758, and the National Science Foundation under NSF Award 1922658. Additionally, we acknowledge support by NSERC (Discovery Grant: RGPIN-2020-05105; Discovery Accelerator Supplement: RGPAS-2020-00031; Arthur B. McDonald Fellowship: 566355-2022) and CIFAR (Canada AI Chair; Learning in Machine and Brains Fellowship). This research was enabled in part by support provided by (Calcul Québec) (https://www.calculquebec.ca/en/) and the Digital Research Alliance of Canada (https://alliancecan.ca/en). The authors acknowledge the material support of NVIDIA in the form of computational resources.
AGC is supported by the Natural Sciences and Engineering Research Council of Canada (NSERC), PGSD3-559278-2021. 
DL is supported by the Natural Sciences and Engineering Research Council of Canada (NSERC) \#CGSD3-569390-2022 and \#CGS-MSFSS-588158-2023. Additionally, M.S. was supported by  Fonds de recherche du Québec – Nature et Technologies (FRQNT) \#2023-2024-B2X-336394 and UNIQUE PhD Excellence Scholarship. We are grateful to Alison Gopnik, Eunice Yiu, and Fei Dai for their generous and insightful feedback on earlier results, which were invaluable in shaping the direction of this project. We are also thankful for insightful discussions with Shannon Yasuda, Solim LeGris, and Jeff Guo which have made this paper better in many ways.

\bibliography{biblio}
\bibliographystyle{colm2025_conference}

\newpage
\appendix
\section*{Appendix}

\section{Related Works}
\label{sec:related-works}

\paragraph{The Blicket-detecting Machine}

The Blicket detector task has a long history in development psychology extending back to seminal studies in causal reasoning and categorization \citep{Gopnik2000, Gopnik2004, Sobel2004, Gopnik2007, Sobel2007, Lucas2014, Walker2014, Bonawitz2010}. 
The core premise of the Blicket detector is a machine that activates under specific conditions, challenging an agent (in the original setting a child) to discover which objects (called ``Blickets'') trigger the machine and according to what rule (also known as ``over-hypothesis''). A disjunctive rule will activate the machine when any Blicket is present, whereas a conjunctive rule requires multiple Blickets to activate the machine. By exploring object combinations and/or observing machine activation, agents must generate and test hypotheses about causal relationships. While simple, it reveals profound insights into how intelligent systems—whether humans or AI—discover and reason about causal structures. In our study, we translate this experimental paradigm into a text-based interaction, probing LM's capacity to explore, infer, and reason about causal relationships.

\paragraph{Reasoning and Exploration in LMs}

Recent advancements in LMs have significantly expanded their applications in autonomous decision-making settings \cite{brown2020language,openai2023gpt4,bubeck2023sparks}. Particularly, LMs have been successfully applied in tasks requiring exploration and reasoning, such as interactive decision-making \cite{yao2022react,shinn2023reflexion,wang2023voyager} and scientific discovery \cite{boiko2023emergent}. For example, Yao et al. (2022) proposed ReAct, enabling LMs to interleave reasoning and acting, thus improving decision-making capabilities \cite{yao2022react}. More recent frameworks like Reflexion \citep{shinn2023reflexion} and Voyager \citep{wang2023voyager} further leverage iterative exploration and reflection to boost autonomous exploration performance.
However, existing literature also points to significant limitations in the exploratory capabilities of LMs. Osband et al. (2016) highlight that effective exploration remains a fundamental challenge even in traditional reinforcement learning agents \citep{osband2016deep}. Recent work by Liu et al. (2023) further underscores the insufficient exploratory behaviours of LMs, suggesting that their tendency to imitate human-generated text leads to suboptimal decision-making strategies \cite{liu2023agentbench}. Our work complements these findings by systematically studying whether LMs exhibit biases specifically in causal inference contexts, thus bridging a gap in current understanding of LM exploration and inference.

A closely related work to our hypothesis sampling agent is \citet{piriyakulkij2024experiment}, which similarly study a Blicket-like environment and samples hypotheses from an LM. Our work is complementary to theirs, though we focus much more on investigating the LM's ability (and failure) to do causal exploration and reasoning, in a task where the LM needs to perform more actions, and evaluate over a greater range of models. 
Another related work is \citet{wang2023hypothesis} which similarly generates code as hypotheses, although focusing on the inductive reasoning ARC task rather than agentic causal discovery tasks.

\paragraph{Biases in Human and Machine Reasoning} Psychological research extensively documents systematic biases in human causal reasoning \cite{tenenbaum2007intuitive, griffiths2009theory, lucas2014children}. Among these, the Blicket paradigm introduced by Gopnik and Sobel (2000) \cite{gopnik2000detecting} has emerged as a canonical method for investigating causal inference in humans, particularly children. Crucially, research using Blicket tests shows a robust disjunctive bias, where humans, especially adults, systematically favour simpler disjunctive causal hypotheses over conjunctive ones, even when evidence supports the latter equally or more strongly \cite{lucas2014children, bonawitz2012children, sobel2004children}. These biases likely arise from human reliance on cognitive heuristics or simplicity priors during causal reasoning

Previous studies indicate that LMs internalize human-like cognitive biases from their training data, including reasoning shortcuts or heuristics \citep{hagendorff2023machine, dasgupta2022language}. Our study directly addresses whether these biases also manifest in LMs performing causal reasoning tasks, using the Blicket paradigm to rigorously quantify such tendencies.

\section{Experimental Details}

\subsection{The Oracle InfoGain Agent}
\label{app:oracle-infogain-description}

For objects' states $x \in \mathcal{X}$, machine state $y \in \mathcal{Y}$, and the set of all hypotheses $f \in \mathcal{F}$, the InfoGain agent maximizes expected information gain $G(x)$ at each step. It iterates through the following: 
\begin{enumerate}
    \item Generating a set of candidate next observations $x$ to reach,
    \item For each candidate, compute $G(x)$,
    \item Go to x that maximizes $G(x)$.
\end{enumerate}

The set of candidate next observations are all results of single actions. We represent it as a binary vector. E.g. if the current state is [0, 1, 1], one candidate next state could be [0, 0, 1], another is [0, 1, 0], and so on. For N objects, there are N such candidates at each step.

Concretely, we have the space of hypotheses $f \in \mathcal{F}$ which is all possible items being the Blicket / not a Blicket, and the two rules (disj / conj). There are $2^{N+1}$ hypotheses total for N objects. Algorithmically we represent them as boolean masks (for which objects are Blicket) and \texttt{numpy} operations. E.g. a conjunctive rule where the last two objects are Blicket is:
\begin{lstlisting}
def func(x):
    mask = np.array([0, 1, 1], dtype=bool)
    return np.all(x[mask])
\end{lstlisting}

For each candidate $x$, we compute:
\begin{equation}
    G(x) = H(F) - \mathbb{E}_{p(y|x)} \big[ H(F | x, y) \big]
\end{equation}
Below we show all components of $G(x)$ are computable quantities. First, $H(F)$ is the prior entropy:
\begin{equation}
    H(F) = - \sum_{f \in \mathcal{F}} p(f) \log p(f)
\end{equation}
where $p(f)$ is uniformly distributed over all $f$'s consistent with all observations so far.

The second term is,
\begin{equation}
    \mathbb{E}_{p(y|x)} \big[ H(F|x,y) \big] = \sum_y p(y|x) \big[ -\sum_f p(f|y,x) \log p(f|y, x)\big]
\end{equation}
Note the summation over $y$ amounts to iterating over all values of $y$, which in our case is binary $y \in \{0, 1\}$ (machine off or on).

Finally, we can write using Bayes,
\begin{equation}
    p(f|y, x) = \frac{p(y|f, x) p(f)}{\sum_f p(y|f, x) p(f)}
\end{equation}
where $p(f)$ is from the prior as defined above. $p(y|f, x) = 1$ if $f(x) = y$, and 0 otherwise.

\subsection{Language Model Access}
\label{app:lm-version-access}

We select a mix of models to span several key axes---reasoning capability, capacity, openness, and architectural family---to explore how each factor may influence hypothesis elimination and exploration behaviour. The \textbf{GPT} family represents some of the highest-performing, state-of-the-art LMs available. Within this family, we include both \texttt{gpt-4o} and \texttt{gpt-4o-mini}, enabling a comparison of model size effects within similar architectures. The \textbf{DeepSeek} family offer a complementary point of comparison: it includes state-of-the-art LMs in their own right, allowing us to compare reasoning model vs. non reasoning model, along with full access to intermediate reasoning content. We also study a few small-to-medium sized models, including \textbf{QwQ} (medium sized reasoning model from Qwen), and the \textbf{gemma3} series of models ($<30$B). 

We accessed the GPT models and DeepSeek models using the OpenAI and DeepSeek APIs, respectively. Specifically, we use \texttt{gpt-4o-2024-08-06} and \texttt{gpt-4o-mini-2024-07-18}. As DeepSeek does not provide reproducible access specifications that we are aware of, we report we access the DeepSeek-V3 (\texttt{deepseek-chat}) and DeepSeek-R1 (\texttt{deepseek-reasoner}) models over the course of March 23 to March 28, 2025. Otherwise, we use Ollama \citep{ollama2025}.

\subsection{Statistical Validation}
\label{app:stat-validation-main-result}

We evaluate all LMs in the ``Blicket Test'' text environment detailed in Section~\ref{sec:the-blicket-test}. For experiments in Section~\ref{sec:overall-perf-results}-\ref{sec:exploration-results} and Section~\ref{sec:hypothesis-sampling-agent}, we run the agent in the game with either 4 or 8 objects, where 2 of them are Blickets (randomly assigned at the start of each episode). All objects start off on the floor or on the machine with 0.9 or 0.1 independent probability. The agent is allowed up to 32 actions in the environment before the game transitions to the Q\&A phase where the agent is asked binary questions about the identity of the Blickets.

For the main results (Figure~\ref{fig:blicket-perf-main-result}), we use a minimum of 16 independent trials per combination of model + prompt variant + system prompt. Specifically for Figure~\ref{fig:blicket-perf-best-lms}, we plot the \textit{best} system message and prompt for \textit{each} model, averaging over random seeds to compute the 95 standard error of mean. 

\subsection{System Message}
\label{sec:system-message-defaults}

We carefully design three different system message variants based on how information is presented to human participants historically in Blicket Test experiments \citep{Lucas2014}. The first is \texttt{human default}, which is based on a common-sense explanation of the task. The second is \texttt{human conjunctive}, which makes use of the description ``Blickets have blicketness''. As \citet{Lucas2014} notes, this induces more conjunctive semantics. Finally, the variant \texttt{math definition} provides a rigorous math definition to describe Blickets which maximally avoid ambiguity.

The general system message prompt template follows the following format:
\begin{lstlisting}
[## INITIAL MESSAGE ##]

Your goals are:
- Identify which objects are blickets.
- Infer the underlying rule for how the machine turns on. 

Here are the available commands:
  look:                describe the current room
  put ... on ...:      put an object on the machine or the floor
  take ... off ...:    take an object off the machine
  exit:                exit the game

[## TIPS ##]

You have #HORIZON# steps to complete the task. You can also exit the task early if you think you understand the relationship between the objects and the machine. After the task is done, you will be asked which objects are blickets, and the rule for turning on the machine.

You will be prompted at each turn to choose actions.
\end{lstlisting}

\subsubsection{Human Default Variant}
Initial Message:
\begin{lstlisting}
You are an intelligent, curious agent. You are playing a game where you are in a room with different objects, and a machine. Some of these objects are blickets. You can't tell which object is a blicket just by looking at it. Blickets make the machine turn on, following some hidden rule.
\end{lstlisting}

Tips:
\begin{lstlisting}
Tips:
- All objects can be either on the machine or on the floor.
- You should think about how to efficiently explore the relationship between the objects and the machine.
\end{lstlisting}

\subsubsection{Human Conjunctive Variant}
Initial Message:
\begin{lstlisting}
You are an intelligent, curious agent. You are playing a game where you are in a room with different objects, and a machine. Some of these objects are blickets. You can't tell which object is a blicket just by looking at it, but they have blicketness inside of them. Blicketness makes the machine turn on, following some hidden rule.
\end{lstlisting}

Tips:
\begin{lstlisting}
Tips:
- All objects can be either on the machine or on the floor.
- You should think about how to efficiently explore the relationship between the objects and the machine.
\end{lstlisting}

\subsubsection{Math Definition  Variant}
Initial Message:
\begin{lstlisting}
You are an intelligent, curious agent. You are playing a game where you are in a room with different objects, and a machine. Some of these objects are blickets. You can't tell which object is a blicket just by looking at it, but they have blicketness inside of them. Blicketness makes the machine turn on, following some hidden rule.
\end{lstlisting}

Tips:
\begin{lstlisting}
Tips:
- All objects can be either on the machine or on the floor.
- You should think about how to efficiently explore the relationship between the objects and the machine.
- A blicket is defined as an object whose state is not independent of the state of the machine (in other words, the object's state distribution and the machine's state distribution have nonzero mutual information)
\end{lstlisting}

\subsection{Prompting Techniques}
\label{app:prompting-techniques}

For non-reasoning models, we also use a number of standard prompting techniques, including ReAct \citep{yao2022react}, Reflexion \citep{shinn2023reflexion}, and Chain-of-Thought (CoT) \citep{wei2022chain}. As the point of the Blicket Test is to discover knowledge, we do not provide it with few-shot examples, but ask the LMs to do exploration zero-shot, based on generic prompts. The prompting technique is applied both during exploration (where agent is prompted to act), and Q\&A (where the agent answers True/False). Details below.

Default Prompt
\begin{lstlisting}
Directly output the command in the format '> command'. Ensure only one command is included.
\end{lstlisting}

ReAct Prompt
\begin{lstlisting}
First briefly reason and think about your plan to solve the task. Then, output the command in the format '> command'. Ensure only one command is included.
\end{lstlisting}

Reflexion Prompt
\begin{lstlisting}
Please first reflect on your plan to solve the task. Explain your reasoning and self-evaluate any potential issues before outputting the final command in the format '> command'. Ensure only one command is included.
\end{lstlisting}

Chain-of-Thought Prompt
\begin{lstlisting}
Please provide a detailed chain-of-thought outlining all intermediate steps leading to your decision, and then output the final command in the format '> command'. Ensure only one command is included.
\end{lstlisting}

\subsection{Example Trajectory}
\label{sec:example-qa-full-trajectory}

Here we provide an example full trajectory in an environment with 3 object, with conjunctive (AND) rule. This is the full prompt given to the agent during the Q\&A phase, which contains the entire exploration interaction history as context. In this environment, objects 1 and 2 are Blickets. 

\begin{lstlisting}
You are in a room. You see a machine at the center of this room. 

There are also 3 objects scattered around the room. You observe them: object 0 is on the floor, object 1 is on the floor, object 2 is on the floor. 

The machine hums softly in front of you, seemingly waiting. The light on the machine is currently off. You wonder if there is a relationship between the objects and the machine.

> put object 0 on machine
You put object 0 on the machine. The light on the machine is currently off.
> put object 1 on machine
You put object 1 on the machine. The light on the machine is currently off.
> put object 2 on machine
You put object 2 on the machine. The light on the machine is now on.
> take object 0 off machine
You took object 0 off of the machine. The light on the machine is now on.
> take object 1 off machine
You took object 1 off of the machine. The light on the machine is currently off.
> put object 1 on machine
You put object 1 on the machine. The light on the machine is now on.
> take object 2 off machine
You took object 2 off of the machine. The light on the machine is currently off.
> take object 1 off machine
You took object 1 off of the machine. The light on the machine is currently off.
> put object 2 on machine
You put object 2 on the machine. The light on the machine is currently off.
> put object 1 on machine
You put object 1 on the machine. The light on the machine is now on.
> exit
Exiting the episode.

Based on the information you have gathered, answer the following question: Is object 2 a blicket?

Directly output the answer in the format '> True/False'. Ensure only one answer is included.
\end{lstlisting}

\subsection{Example intermediate tokens during Q\&A}

We provide qualitative examples of intermediate tokens below. While intermediate tokens have been used as a proxy for ``reasoning'' in the literature, we do not wish to draw strong conclusions here. Recent evidence suggests their role in the LM’s computation is illusive, for instance, they may not necessarily have to be correct for the final answer to be correct \citep{stechly2025beyond}.

\subsubsection{GPT-4o, react prompting (correct answer)}
Q: Is object 1 a blicket? (true blicket indices: [1, 4])

\begin{lstlisting}
From the observations, the machine turns on when object 1 and object 4 are on the machine together. When object 1 is on the machine alone, the light is off. When object 4 is on the machine alone, the light is off. However, when both objects 1 and 4 are on the machine, the light turns on. This suggests that the rule for the machine to turn on requires both objects 1 and 4 to be present, indicating that both are blickets.

Therefore, object 1 is a blicket.

> True
\end{lstlisting}

\subsubsection{GPT-4o, react prompting (incorrect answer)}
Q: Is object 5 a blicket? (true blicket indices: [2, 5])

\begin{lstlisting}
From the observations, we can see that:

- When object 5 is on the machine alone, the light is off.
- When object 5 is on the machine with object 2, the light is on.
- When object 2 is on the machine alone, the light is on. [incorrect]

This suggests that object 2 is a blicket because it can turn the machine on by itself. Object 5 does not turn the machine on by itself, indicating it is not a blicket. The machine turns on when object 2 is present, regardless of whether object 5 is also present.

Therefore, object 5 is not a blicket.

> False
\end{lstlisting}

\subsubsection{DeepSeek-R1 (correct answer)}
Q: Is object 2 a blicket? (true blickets: [2, 3] conjunctive)
\begin{lstlisting}
Okay, let's try to figure out if object 2 is a blicket. So, the machine turns on when certain objects (blickets) are placed on it. From the steps taken, let's recap what happened.

First, when objects 5 and 0 were placed on the machine, the light stayed off. Then adding 1, 2, and 3 caused the light to turn on. When objects 5 and 1 were removed, the light stayed on. But when object 3 was removed, the light turned off. Then removing 2 also left it off. 

[omitted ~1k characters]

Then removing 3 turned it off. Then putting 3 back alone didn't help, but adding 2 again made it work. So yes, 2 and 3 together activate the machine. Therefore, object 2 is a blicket because when combined with 3, the machine turns on. So the answer should be True.
\end{lstlisting}

\subsubsection{DeepSeek-R1 (incorrect answer)}
Q: Is object 5 a blicket? (true blickets: [5, 7] conjunctive)
\begin{lstlisting}
Okay, let's try to figure out if object 5 is a blicket. From the actions taken in the game, here's what happened:

Initially, all objects except object 0 were on the floor, and the machine was off. The user started by putting objects 1 through 7 one by one onto the machine. Each time, the machine's light stayed off until object 7 was added. When all objects (1-7) were on the machine, the light turned on. Then, removing object 7 caused the light to turn off, and putting it back on turned it on again.

This suggests that the machine's rule requires the presence of object 7 to activate. Since adding all other objects (1-6) didn't trigger the light until 7 was added, it's likely that object 7 is a blicket. However, the other objects (including 5) were on the machine both when the light was off (before 7 was added) and when it was on (after 7 was added). Since the presence of 5 alone (without 7) didn't activate the machine, and removing 7 deactivates it regardless of 5's presence, it seems that 5 isn't a blicket. The rule might be that the machine only needs at least one blicket, which is 7 here. Therefore, object 5 is not a blicket.
\end{lstlisting}

\subsection{Human inference experiments}
\label{app-human-inference}

To study whether LMs are biased by prior assumptions during inference, we conducted a direct textual replication of the inference experiment used in \cite{lucas2014children} and \cite{gopnik2017changes}, designed to probe causal reasoning under uncertainty. First, in the ``training'' phase, the participants (in our case, LMs) are presented with trials that provide evidence suggesting one of three scenarios: (1) A disjunctive causal rule governs the machine; (2) a conjunctive causal rule governs the machine; (3) the evidence is ambiguous and insufficient to definitively determine which causal structure governs the machine (the ``baseline'' condition). Below is the full trajectory of one training trial for each of the 3 scenarios:
(1) disjunctive evidence:
\begin{lstlisting}
You are in a room. You see a machine at the center of this room.

There are also 3 objects scattered around the room. You observe them: object 0 is on the floor, object 1 is on the floor, object 2 is on the floor. The machine hums softly in front of you, seemingly waiting. The light on the machine is currently off. You wonder if there is a relationship between the objects and the machine.
> put object 0 on machine
You put object 0 on the machine. The light on the machine is now on.
> take object 0 off machine
You took object 0 off the machine. The light on the machine is currently off.
> put object 1 on machine
You put object 1 on the machine. The light on the machine is currently off.
> take object 1 off machine
You took object 1 off the machine. The light on the machine is currently off.
> put object 2 on machine
You put object 2 on the machine. The light on the machine is now on.
> take object 2 off machine
You took object 2 off the machine. The light on the machine is currently off.
> put object 0 on machine
You put object 0 on the machine. The light on the machine is now on.
> put object 1 on machine
You put object 1 on the machine. The light on the machine is now on.
> take object 1 off machine
You took object 1 off the machine. The light on the machine is now on.
> put object 2 on machine
You put object 2 on the machine. The light on the machine is now on.
> take object 0 off machine
You took object 0 off the machine. The light on the machine is now on.
> put object 1 on machine
You put object 1 on the machine. The light on the machine is now on.
\end{lstlisting}
(2) conjunctive evidence:
\begin{lstlisting}
You are in a room. You see a machine at the center of this room.

There are also 3 objects scattered around the room. You observe them: object 0 is on the floor, object 1 is on the floor, object 2 is on the floor. The machine hums softly in front of you, seemingly waiting. The light on the machine is currently off. You wonder if there is a relationship between the objects and the machine.

> put object 0 on machine
You put object 0 on the machine. The light on the machine is currently off.
> take object 0 off machine
You took object 0 off the machine. The light on the machine is currently off.
> put object 1 on machine
You put object 1 on the machine. The light on the machine is currently off.
> take object 1 off machine
You took object 1 off the machine. The light on the machine is currently off.
> put object 2 on machine
You put object 2 on the machine. The light on the machine is currently off.
> take object 2 off machine
You took object 2 off the machine. The light on the machine is currently off.
> put object 0 on machine
You put object 0 on the machine. The light on the machine is currently off.
> put object 1 on machine
You put object 1 on the machine. The light on the machine is currently off.
> take object 1 off machine
You took object 1 off the machine. The light on the machine is currently off.
> put object 2 on machine
You put object 2 on the machine. The light on the machine is now on.
> take object 0 off machine
You took object 0 off the machine. The light on the machine is currently off.
> put object 1 on machine
You put object 1 on the machine. The light on the machine is currently off.
\end{lstlisting}
(3) ambiguous evidence: 
\begin{lstlisting}
You are in a room. You see a machine at the center of this room.

There are also 3 objects scattered around the room. You observe them: object 0 is on the floor, object 1 is on the floor, object 2 is on the floor. The machine hums softly in front of you, seemingly waiting. The light on the machine is currently off. You wonder if there is a relationship between the objects and the machine.

> put object 0 on machine
You put object 0 on the machine. The light on the machine is currently off.
> take object 0 off machine
You took object 0 off the machine. The light on the machine is currently off.
> put object 0 on machine
You put object 0 on the machine. The light on the machine is currently off.
> take object 0 off machine
You took object 0 off the machine. The light on the machine is currently off.
> put object 1 on machine
You put object 1 on the machine. The light on the machine is currently off.
> take object 1 off machine
You took object 1 off the machine. The light on the machine is currently off.
> put object 1 on machine
You put object 1 on the machine. The light on the machine is currently off.
> take object 1 off machine
You took object 1 off the machine. The light on the machine is currently off.
> put object 1 on machine
You put object 1 on the machine. The light on the machine is currently off.
> take object 1 off machine
You took object 1 off the machine. The light on the machine is currently off.
> put object 0 on machine
You put object 0 on the machine. The light on the machine is currently off.
> put object 2 on machine
You put object 2 on the machine. The light on the machine is now on.
\end{lstlisting}

Next, in the ``test phase'', the participants (or LMs) are presented with a test trial with new objects and the same machine governed by the causal rule introduced in the training trials. Below is the full trajectory for the test trial:
\begin{lstlisting}
You are in a new room. You see the same machine as the one you previously saw at the center of this room.

You now have 3 different objects scattered around the room. You observe them: object A is on the floor, object B is on the floor, object C is on the floor. The machine hums softly in front of you, seemingly waiting. The light on the machine is currently off. You wonder if there is a relationship between the objects and the machine.

> put object A on machine
You put object A on the machine. The light on the machine is currently off.
> take object A off machine
You took object A off the machine. The light on the machine is currently off.
> put object A on machine
You put object A on the machine. The light on the machine is currently off.
> take object A off machine
You took object A off the machine. The light on the machine is currently off.
> put object A on machine
You put object A on the machine. The light on the machine is currently off.
> take object A off machine
You took object A off the machine. The light on the machine is currently off.
> put object B on machine
You put object B on the machine. The light on the machine is currently off.
> take object B off machine
You took object B off the machine. The light on the machine is currently off.
> put object A on machine
You put object A on the machine. The light on the machine is currently off.
> put object C on machine
You put object C on the machine. The light on the machine is now on.
> put object B on machine
You put object B on the machine. The light on the machine is now on.
> take object B off machine
You took object B off the machine. The light on the machine is now on.

Based on the information above, is object A a blicket?

Directly output the answer in the format '> True/False'. Ensure only one answer is included.

\end{lstlisting}

The test trial is intentionally designed such that the ground truth can be one of the two possibilities: either both A and C are Blickets, and the machine follows a conjunctive rule, or only C is a Blicket, and the machine follows a disjunctive rule.  Participants (or LMs) are then asked whether each test object (A, B, C) is a Blicket. In particular, the answer to the identity of the test object A as a potential Blicket (causal agent) depends on the what the participant (or LM) believes the underlying rule of the causal system is. As such, we quantify the model's propensity to classify A as a Blicket, which serves as a proxy for identifying cognitive bias. The experimental setup mirrors the original psychological study, translated into a text-based interaction format that allows LMs to reason about causal relationships.

\section{Additional Results}

\subsection{Causal Exploration Progress}
\label{app:causal-exploration-progress}

\begin{figure}[h]
    \centering
    \begin{subfigure}[b]{0.47\textwidth}
        \centering
        \includegraphics[width=\textwidth]{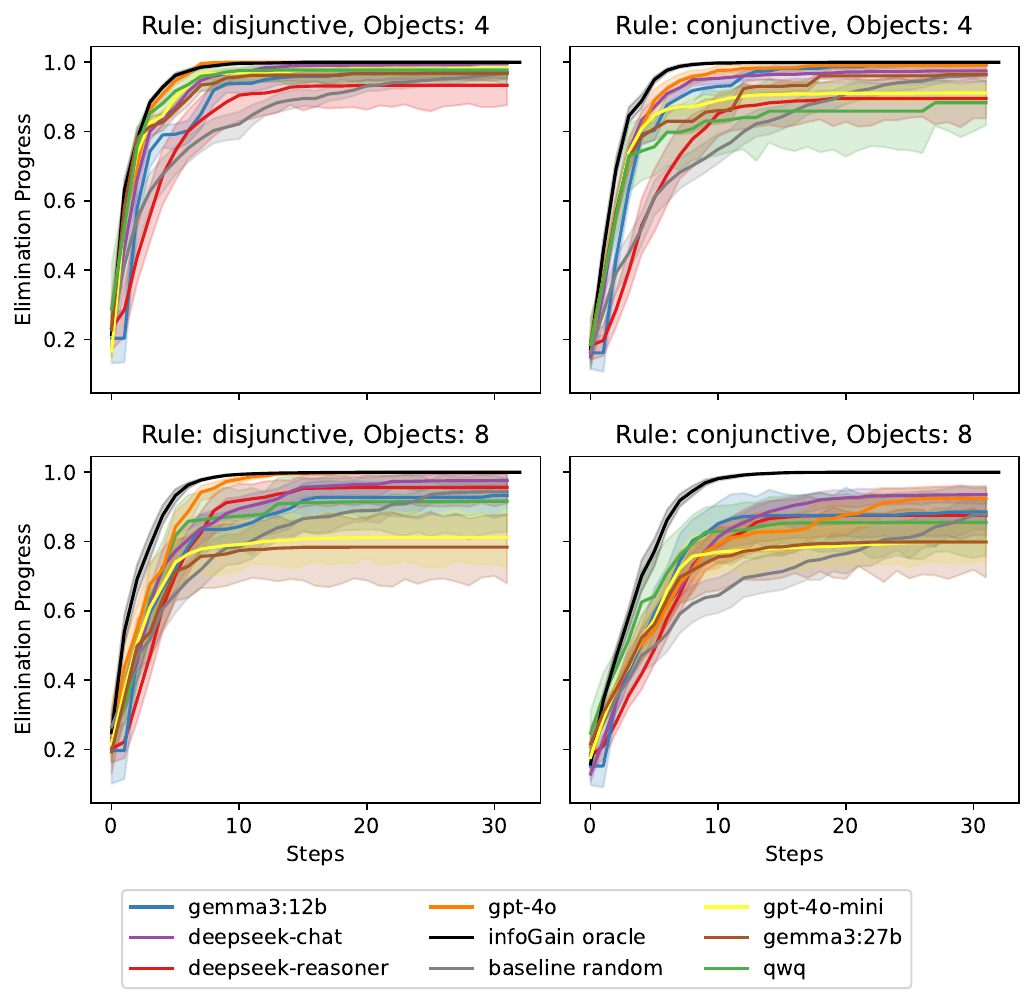}
        \caption{Hypothesis elimination progress, measured between 0 (no hypotheses eliminated) and 1 (all but one eliminated). Higher is better.}
        \label{fig:hypothesis-elim-progress}
    \end{subfigure}
    \hfill
    \begin{subfigure}[b]{0.47\textwidth}
        \centering
        \includegraphics[width=\textwidth]{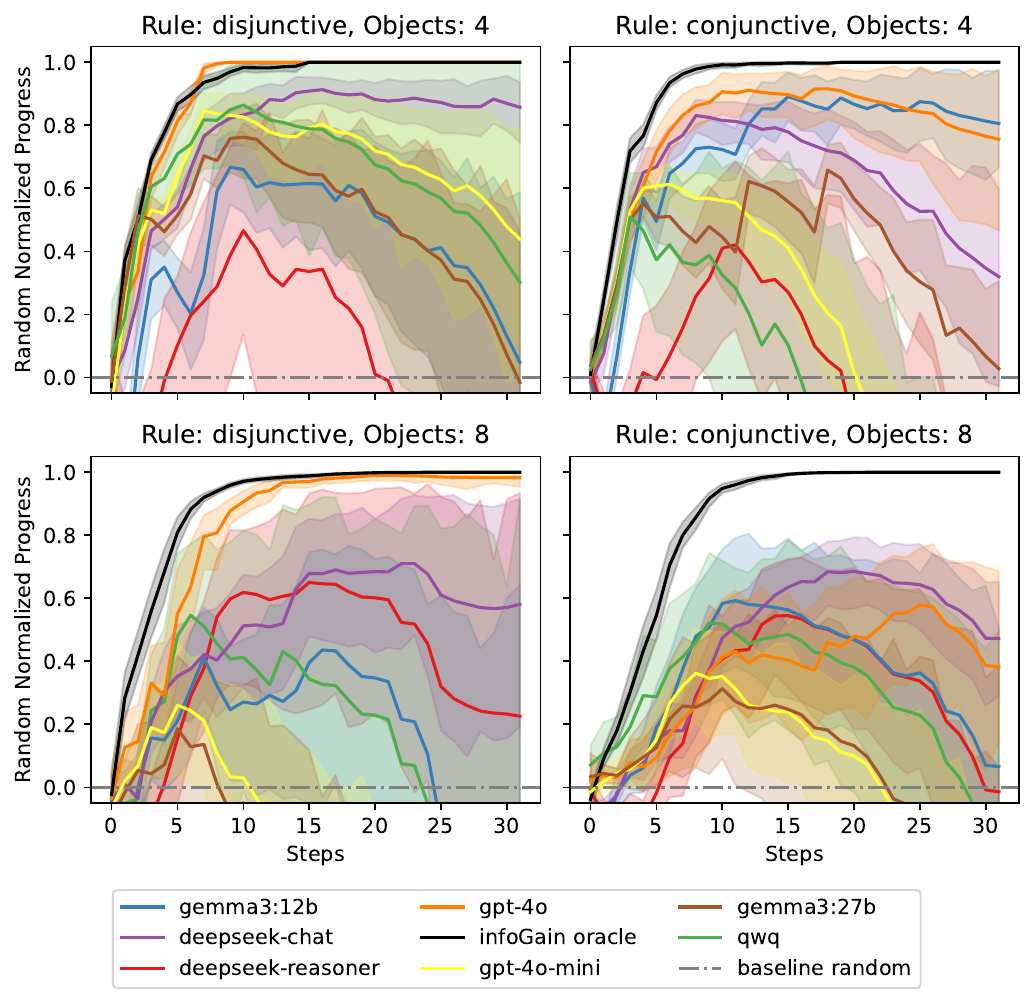}
        \caption{Hypothesis elimination progress normalized by the baseline random agent's mean progress. Normalized progress can be negative, see Equation~\ref{eq:normalized-exploration-progress}.}
        \label{fig:hypothesis-elim-progress-normalized}
    \end{subfigure}
    \caption{Raw and normalized hypothesis elimination progress. Error bars denote standard error.}
\end{figure}

For more comprehensive interpretations of each model's exploration efficiencies, we plot additional visualizations of the results in Figure~\ref{fig:hypothesis-elim-explore-steps}. Figure~\ref{fig:hypothesis-elim-progress} plots the hypotheses elimination as a progress between 0 and 1. Denote the total number of hypotheses as $N$, and number of hypotheses model \texttt{m} has eliminated at time $t$ as $n_{\text{m}}(t)$. The hypotheses elimination progress $\rho$ is measured as,
\begin{equation}
    \rho_\text{m}(t) = \frac{N - n_{\text{m}}(t)}{N - 1} \,,
\end{equation}
which goes to $1$ when all but one hypotheses are eliminated. 

We also consider the possibility that due to bounded computational resources, one task may be harder. We account for this potential difference in search difficulties by normalizing the performance relative to a baseline that takes random actions in each environment (Figure~\ref{fig:hypothesis-elim-progress-normalized}). This normalized progress is measured by,
\begin{equation}
    \bar{\rho}_\text{m}(t) = \frac{\rho_{\text{m}}(t) - \rho_{\text{random}}(t)}{1 - \rho_{\text{random}}(t)} \,,
    \label{eq:normalized-exploration-progress}
\end{equation}
which shows the progress of a model relative to progress made by the random action baseline. Intuitively, the normalized progress measures where the model progress $\rho_{\text{m}}(t)$ is between the random baseline's progress $\rho_{\text{random}}(t)$  and 1. Note that this progress can be negative if the progress a model makes is slower than that of a baseline. We further observe that for the best performing models, LM agents still show a disjunctive bias in exploration efficiencies when normalized by the random action baseline. For completeness, we also report the end-of-episode progress of the performant models in Table~\ref{tab:abs-norm-explore-progress}.

\begin{table}[ht]
\centering
\small
    \begin{tabular}{c|c|cc|cc}
    \toprule
    \textbf{\# Objs} & \textbf{Model} & \textbf{Disj.} & \textbf{Conj.} & \textbf{Norm. Disj.} & \textbf{Norm. Conj.} \\
    \midrule
    \multirow{6}{*}{4} & baseline random   & 0.968 ± 0.049            & 0.964 ± 0.068            & 0                     & 0                     \\
                       & infoGain oracle   & 1.000 ± 0.000            & 1.000 ± 0.000            & 1.000 ± 0.000         & 1.000 ± 0.000         \\
                       & gpt-4o            & \textbf{1.000 ± 0.000}   & 0.991 ± 0.033            & \textbf{1.000 ± 0.000} & 0.756 ± 0.900         \\
                       & gpt-4o-mini       & \textbf{0.982 ± 0.073}   & 0.911 ± 0.113            & \textbf{0.439 ± 2.297} & -1.444 ± 3.109        \\
                       & deepseek-chat     & \textbf{0.995 ± 0.011}   & 0.975 ± 0.046            & \textbf{0.857 ± 0.356} & 0.319 ± 1.255         \\
                       & deepseek-reasoner & \textbf{0.934 ± 0.224}   & 0.895 ± 0.236            & \textbf{-1.085 ± 7.063} & -1.881 ± 6.509       \\
    \midrule
    \multirow{6}{*}{8} & baseline random   & 0.944 ± 0.098            & 0.877 ± 0.183            & 0                     & 0                     \\
                       & infoGain oracle   & 1.000 ± 0.000            & 1.000 ± 0.000            & 1.000 ± 0.000         & 1.000 ± 0.000         \\
                       & gpt-4o            & \textbf{0.999 ± 0.005}   & 0.924 ± 0.125            & \textbf{0.983 ± 0.085} & 0.383 ± 1.019         \\
                       & gpt-4o-mini       & \textbf{0.812 ± 0.357}   & 0.798 ± 0.307            & -2.342 ± 6.346        & \textbf{-0.647 ± 2.499} \\
                       & deepseek-chat     & \textbf{0.976 ± 0.092}   & 0.935 ± 0.138            & \textbf{0.581 ± 1.634} & 0.472 ± 1.121         \\
                       & deepseek-reasoner & \textbf{0.956 ± 0.174}   & 0.875 ± 0.215            & \textbf{0.226 ± 3.091} & -0.014 ± 1.752        \\
    \bottomrule
    \end{tabular}
    \caption{Absolute and normalized exploration progress (± stdev) across models and object counts, at the end of each of episode. \texttt{Disj} and \texttt{Conj} denote absolute progress, while \texttt{Norm Disj} and \texttt{Norm Conj} denote random-normalized progress. Higher is better.}
    \label{tab:abs-norm-explore-progress}
\end{table}

\begin{figure}[h]
  \centering
  \includegraphics[width=0.8\textwidth]{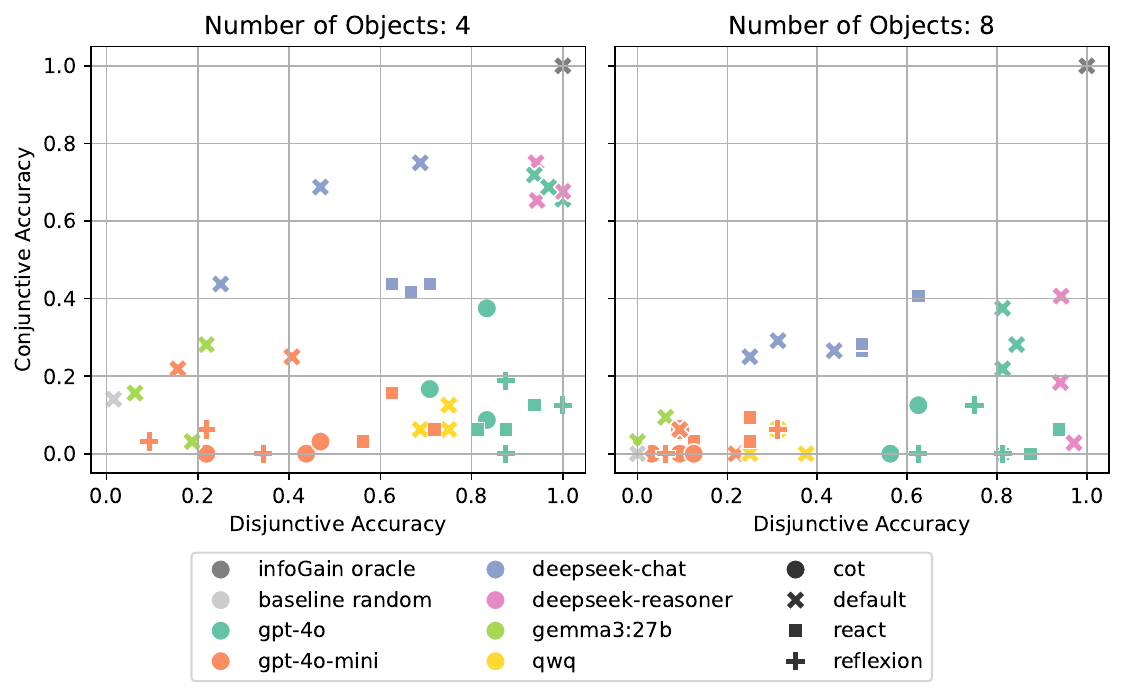} %
  \caption{All models' disjunctive and conjunctive accuracies. Colours indicate model type, while point shapes indicate the prompt. This is an extension of the result in Figure~\ref{fig:blicket-perf-all-models}.}
  \label{fig:dis-conj-acc-4-8-objs}
\end{figure}

\subsection{Effect of reasoning effort on exploration}

\begin{figure}[h]
    \centering
    \begin{subfigure}[b]{0.49\textwidth}
        \centering
        \includegraphics[width=\textwidth]{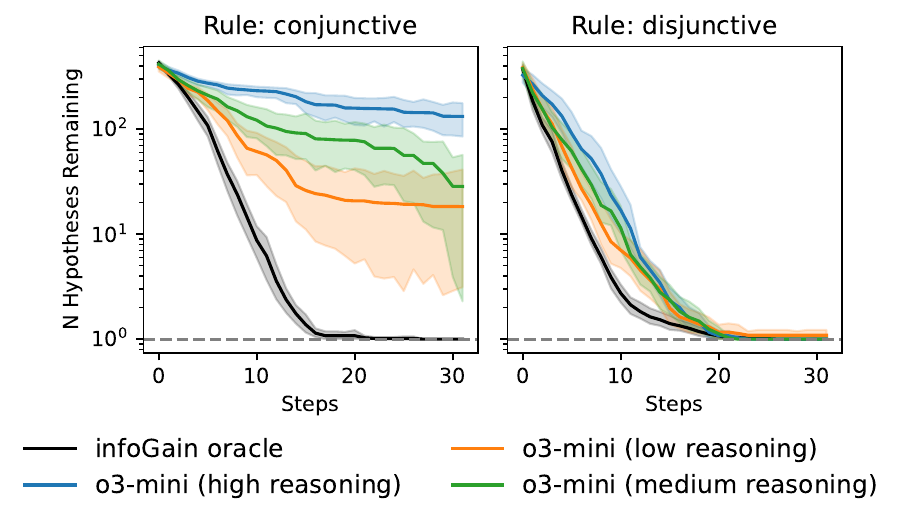}
        \caption{Reasoning model \texttt{o3-mini-2025-01-31}.}
    \end{subfigure}
    \hfill
    \begin{subfigure}[b]{0.49\textwidth}
        \centering
        \includegraphics[width=\textwidth]{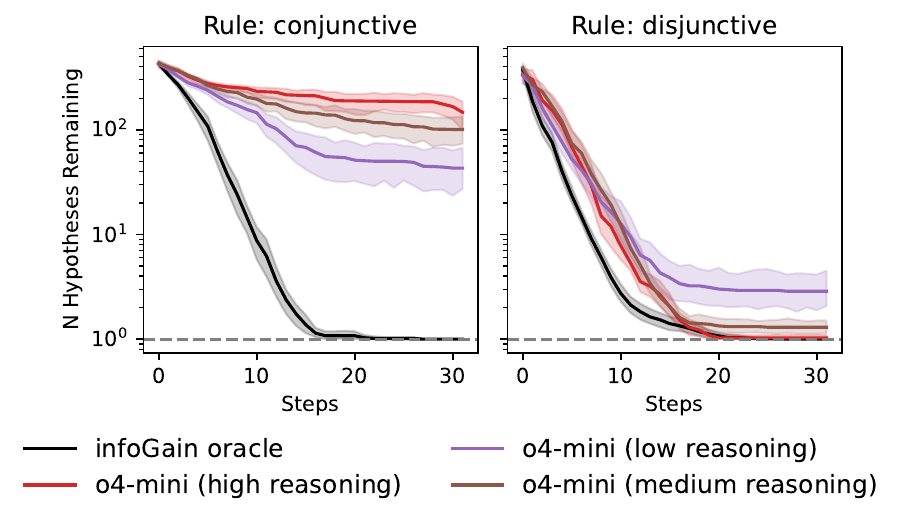}
        \caption{Reasoning model \texttt{o4-mini-2025-04-16}.}
    \end{subfigure}
    \caption{Hypotheses elimination efficiency for the GPT-o-mini series of reasoning models at varying reasoning efforts. Lower is better. Error bar denotes standard error of mean.}
    \label{fig:hypothesis-elim-gpt-o-reasoning}
\end{figure}

To investigate the effect of ``reasoning'' on the models' abilities to explore, we run additional evaluation using the GPT-o-mini series of reasoning models on the 8-objects environments. We report the results in Figure~\ref{fig:hypothesis-elim-gpt-o-reasoning}. In the conjunctive setting (left panels), models with higher reasoning effort consistently lag behind both lower-reasoning variants and the infoGain oracle. We hypothesize this could be due to: (1) the reasoning models evaluated here were overfitted to the verifiable tasks they were trained on; (2) reasoning amplifies existing priors in these models that favours disjunctive interpretations. Future works are needed to address these open questions.

\vspace{-1em}

\subsection{Inference on causal relationships}
\label{app:inference-on-causal-rela-additional}

We report the trajectory statistics of the exploration data used in Section~\ref{sec:inference-on-causal-relationship} in Table~\ref{tab:inference-causal-traj-lengths}.

\begin{table}[h]
    \centering
    \begin{tabular}{lll}
    \toprule
    \textbf{Data Source} & \textbf{Rule Type} & \textbf{Traj Length (± Stdev)} \\
    \midrule
    LLM (GPT-4o)    & Conjunctive   & 18.94 ± 5.04  \\
    LLM (GPT-4o)    & Disjunctive   & 18.06 ± 2.35  \\
    \hline
    Oracle          & Conjunctive   & 12.70 ± 4.79  \\
    Oracle          & Disjunctive   & 14.58 ± 4.28  \\
    \hline
    Object Count    & Conjunctive   & 32.00 ± 0.00  \\
    Object Count    & Disjunctive   & 32.00 ± 0.00  \\
    \hline
    Random          & Conjunctive   & 32.00 ± 0.00  \\
    Random          & Disjunctive   & 32.00 ± 0.00  \\
    \bottomrule
    \end{tabular}
    \caption{Trajectory lengths (mean ± standard deviation) across exploration data sources and rule types. Used in Figure~\ref{fig:data-standarized-inference-perf}.}
    \label{tab:inference-causal-traj-lengths}
\end{table}

\subsection{Human exploration characteristics}
\label{appendix:human-exploration}
\begin{figure}[h]
    \centering
    \includegraphics[width=0.8\linewidth]{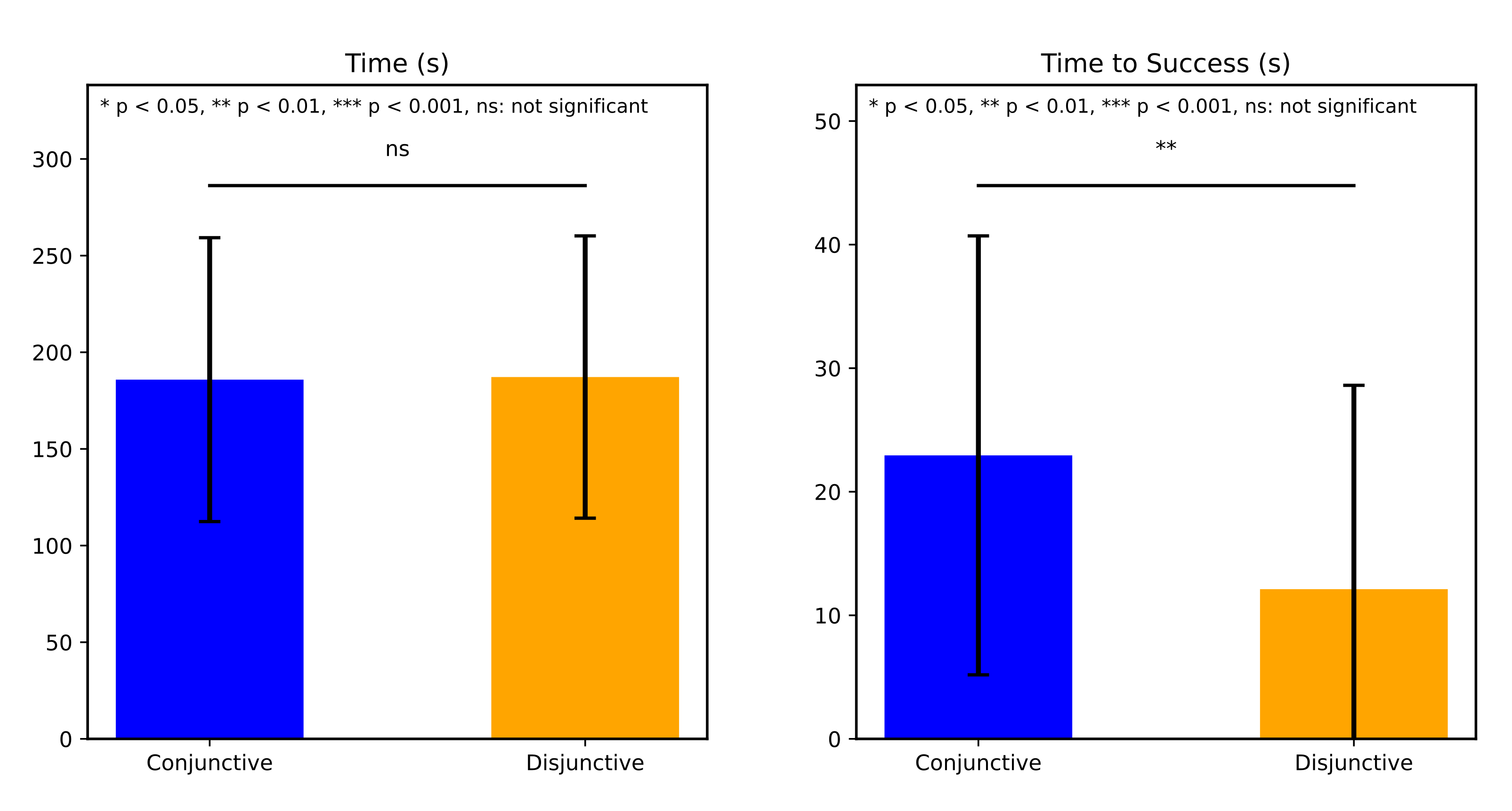}
    \caption{Additional exploration statistics in children, reproduced from Table 1 of \cite{kosoy2022learning} aggregating ``given hypothesis'' and ``not given hypothesis'' conditions. Left: Number of times children checked if state of the Blicket-detecting machine. Right: Average time children played before seeing the Blicket detector go on for the first time. Error bar denote standard deviation. Two-sample t-test significance level denote at the top of each panel.}
    \label{fig:app-child-explore-kosoy-agg}
\end{figure}
\begin{figure}[h]
    \centering
    \includegraphics[width=0.75\linewidth]{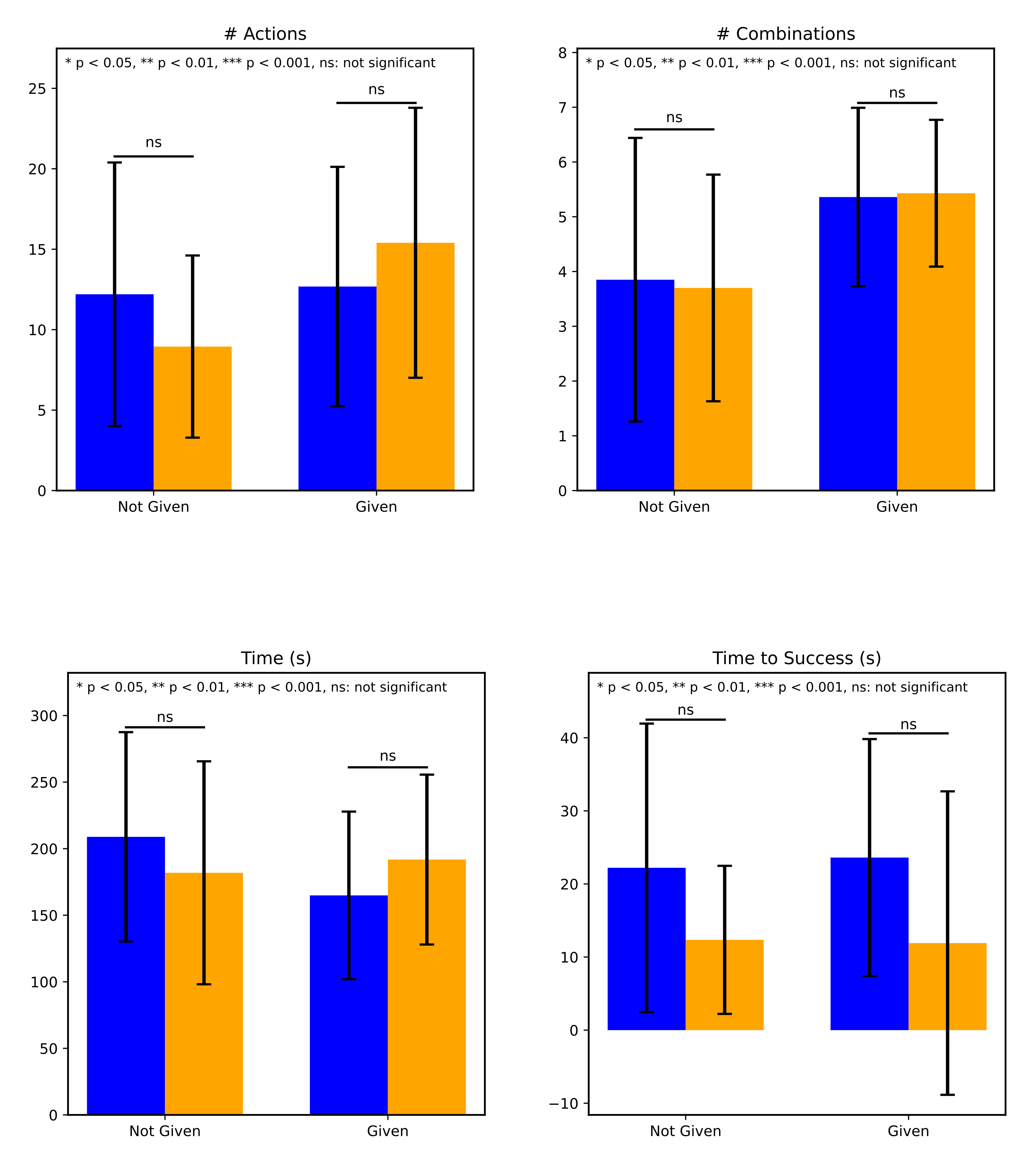}
    \caption{Additional exploration statistics in children, reproduced from Table 1 of \cite{kosoy2022learning} without aggregating.  Top left: Average number of times children pressed the ``check'' button per trial. Top right: Average number of unique combinations attempted per trial. Bottom left: Average total time spent exploring the Blicket-detecting machine. Bottom right: Average time before successfully activating the machine for the first time in a trial. Error bars denote standard deviation. Two-sample t-test significance level denote at the top of each panel.}
    \label{fig:app-child-explore-kosoy}
\end{figure}

\cite{kosoy2022towards} conducted Blicket Test experiment with 4-year-old children and allowed them to freely explore different combinations of given objects to ``figure out how to make the machine go''. Here we reproduce their results from Table 1 graphically and ran statistical tests to have a better understanding of how well LMs explore in comparison to human children \ref{section:lm-explore-bias-children}. In summary, across different experimental conditions (whether the causal rule of the Blicket-detecting machine can be inferred from example trials, i.e. ``given hypothesis'', vs. when the causal rule cannot be inferred, i.e. ``not given hypothesis'') and measurements (number of unique combinations attempted, number of ``checks'' which we use as the equivalent for the number of actions taken before exiting, time spent exploring), children explore similarly for conjunctive machine and disjunctive machine (Figures \ref{fig:app-child-explore-kosoy-agg}, \ref{fig:app-child-explore-kosoy}). 

\newpage
\subsection{Hypothesis Sampling Agent}

\subsubsection{Details of Procedure}
\label{app:hypotheses-sampling-agent-details}

The sampling agent approximates the InfoGain agent to reduce the entropy of the hypothesis distribution. Instead of iterating over candidate next observations and computing $G(x)$, we simply prompt the LM to take actions that minimizes $H(F)$. The procedure is outlined below.

\paragraph{1. Hypotheses sampling to form set of active hypotheses}

If there are zero hypotheses: we prompt the agent to generate hypotheses as python functions.
\begin{lstlisting}
You have seen the following observations so far:
[HISTORICAL OBSERVATIONS]

Come up with some hypothesis about how the world works. Write each one as a python function. The function should:
- Take as input a numpy boolean vector of length #NUM_OBJECTS#, where 0 means object is on floor, and 1 means object is on machine
- Output a boolean, where 0 means machine light is off, and 1 means machine light is on
- First construct a boolean mask indicating which object(s) are blickets, extract just the state of these objects, then define some function over these states to produce the state of the machine.
- Only consider hypothesis where one or all objects that are blickets needs to be put on the machine.
- Do not consider the "sum" operation. Only consider boolean operations.

Produce #NUM_HYPOTHESES# hypothesis, do not generate the same hypothesis. Be diverse in both the blicket masks and the potential functions.

Do not generate the hypothesis you have already eliminated: 
[ELIMINATED HYPOTHESES]

Do not generate hypothesis you have already generated: 
[ACTIVE HYPOTHESES]

Directly output the answer, ensure only one answer is included. Return a set of hypotheses as python functions.
\end{lstlisting}

We save all \textit{valid} (python executable) hypotheses as string python code in the list of currently active hypotheses.

\paragraph{2. Eliminate hypotheses based on observation history}

For each python function $f$ (i.e. hypothesis) in the currently active list, we check if it perfectly predicts all past observations: i.e. $f(x_i) = y_i$ for all $\{(x_i, y_i) \}_{i=0, ..., t}$. If it does not, we move it to the ``eliminated hypotheses'' list.

\paragraph{3. Prompt agent to take actions that eliminate more hypotheses}
We prompt the agent to take actions conditioned on the observations and the list of active hypotheses:
\begin{lstlisting}
You are currently entertaining the following list of hypothesis: 
[ACTIVE HYPOTHESES]

You have seen the following observations so far:
[OBSERVATIONS SO FAR] 

Given the observations so far, and the list of hypotheses (hypothesis space), take an action which will disprove the existing hypothesis.

Directly output the answer, ensure only one answer is included. Output the action in the format \'> action\'. Ensure only one action is included.
\end{lstlisting}

The agent then takes the next actions and we repeat the above procedure for exploration.

\paragraph{Q\&A Phase}
During the Q\&A phase, we prompt the agent to answer questions conditioning on the remaining set of hypotheses:
\begin{lstlisting}
You have seen the following observations so far:
[HISTORICAL OBSERVATIONS]

You have disproven the following hypothesis: 
[ELIMINATED HYPOTHESES]

You have not yet disproven the following hypothesis: 
[ACTIVE HYPOTHESES]

Based on the information above, answer the following question: [QUESTION]

Output the answer in the format \'> True/False\'. Ensure only one answer is included.
\end{lstlisting}

In this way, the sampling agent explicitly represents its beliefs $p(F)$ as a set of active hypotheses, which the LM conditions on to take the most informative action.

\subsubsection{Proof of Remark}
\begin{remark}
    \label{prop:ent-increase}
    Let $p(F)$ be a discrete prior distribution, and we sample \textit{without replacement} from $p(F)$ to construct empirical distribution $q(F)$. Let $q_t(F)$ be the empirical distribution after $t$ unique samples, the entropy of $q$ monotonically increases with each additional unique sample, $H(q_t) < H(q_{t+1})$. 
\end{remark}
\begin{proof}
    Given the already sampled set $S_t = \{f_1, f_2, ..., f_t\}$, the empirical distribution is defined as:
    \begin{equation}
        q_t(f) =\begin{cases}
            \frac{1}{t}   & \text{if } f \in S_t \,, \\
            0   & \text{otherwise}\,.
        \end{cases}
    \end{equation}
    The entropy is $H(q_t) = -\mathbb{E}_{q_t}[\log q_t] = -\log (\frac{1}{t}) = \log t$. $\log$ is a monotonically increasing function in $t$.
\end{proof}

\subsubsection{Exploration for Both Rules}
\begin{figure}[h]
  \centering
  \includegraphics[width=0.8\textwidth]{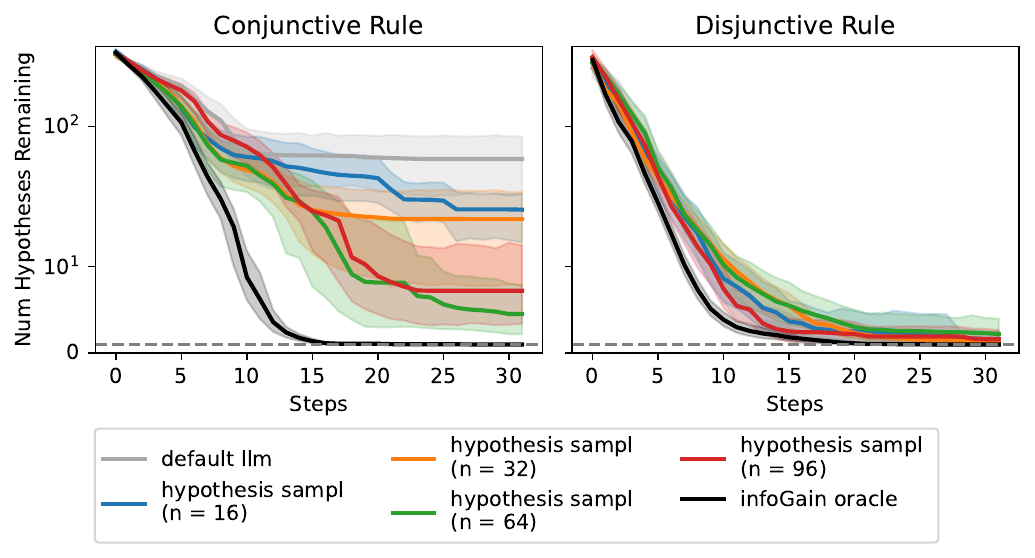} %
  \caption{Hypothesis elimination performance of the hypothesis sampling agent for 8 objects for both disjunctive (``OR'') and conjunctive (``AND'') rules.}
  \label{fig:hypo-elim-8-obj-both-rules}
\end{figure}

See Figure~\ref{fig:hypo-elim-8-obj-both-rules}.

\subsubsection{Hypothesis sampling agent resolves biased exploration patterns}
\label{app:hypothesis-sampl-agent-exploration-patterns}

\begin{figure}[h]
    \centering
    \includegraphics[width=0.7\linewidth]{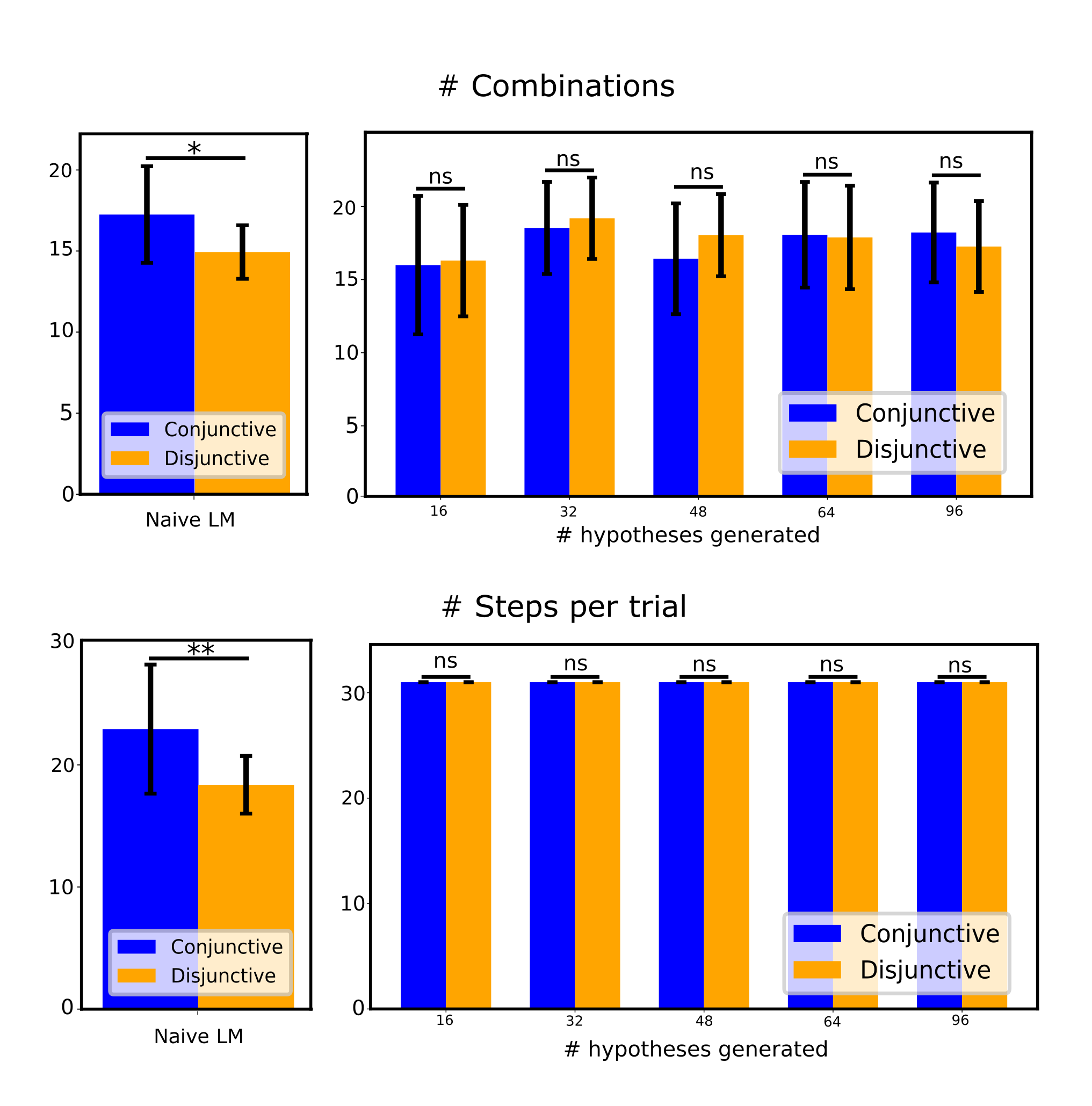}
    \caption{Amount of exploration by gpt-4o in the 8-object Blicket environment with default system prompts, measured by the number of unique combinations of objects attempted per trial (top) and number of steps taken before exiting the trial, for agents before (left) and after (right) inference-time hypothesis sampling, averaged across trials, error bar denote standard deviation. Two-sample t-test: * $p < 0.05$, ** $p < 0.01$, *** $p < 0.001$, ns: not significant.}
    \label{fig:hypo-gen-exploration}
\end{figure}

Additionally, we evaluated whether our method of hypothesis sampling addressed the differences in exploration patterns between conjunctive and disjunctive causal structures that we observed in \ref{section:lm-explore-bias-children}, wherein LMs explore more extensively in the conjunctive condition than disjunctive. We observed that sampling hypothesis significantly reduced the difference in exploration patterns between conjunctive and disjunctive conditions (Figure \ref{fig:hypo-gen-exploration}), suggesting that our method helps LM explore more systematically, much like a curious child.

\end{document}